%% file: elsarticle-template-num.tex
\documentclass[preprint,times,review,10pt]{elsarticle}

\usepackage{amssymb}
\usepackage{amsmath}

\usepackage{float, booktabs, multirow, amssymb, enumitem, colortbl}
\definecolor{mycolor}{rgb}{0.886, 0.949, 0.996}

\newcommand{\revise}[1]{\textcolor{black}{#1}}

\newcommand{\secondrevise}[1]{\textcolor{black}{#1}}

\journal{Nuclear Physics B}

\begin{document}

\begin{frontmatter}

\title{Temporal Grounding of Video Large Language Model via Holistic Event Perception and Matching}

\author[ntu]{Jiahao Nie\fnref{equal}}
\ead{jiahao007@e.ntu.edu.sg}
\author[xjtu]{Wenbin An\fnref{equal}}
\author[damo]{Gongjie Zhang}
\author[isct]{Yicheng Xu}
\author[vinuni,ntu]{Yap-Peng~Tan}
\author[smbu]{Alex~C.~Kot}
\author[ntu]{Shijian Lu\corref{cor}}
\ead{shijian.lu@ntu.edu.sg}

\affiliation[ntu]{organization={Nanyang Technological University},
            addressline={50 Nanyang Avenue}, 
            city={Singapore},
            postcode={639798}, 
            country={Singapore}}
\affiliation[xjtu]{organization={Xi'an Jiaotong University},
            addressline={No.28 Xianning West Road}, 
            city={Xi'an},
            postcode={710049}, 
            country={China}}
\affiliation[damo]{organization={Alibaba DAMO Academy},
            addressline={1 Raffles Place}, 
            city={Singapore},
            postcode={048615}, 
            country={Singapore}}
\affiliation[isct]{organization={Institute of Science Tokyo},
            addressline={2 Chome-12-1 Ookayama}, 
            city={Tokyo},
            postcode={152-8550}, 
            country={Japan}}
\affiliation[vinuni]{organization={VinUniversity},
            addressline={Vinhomes Ocean Park}, 
            city={Hanoi},
            postcode={100000}, 
            country={Vietnam}}
\affiliation[smbu]{organization={SMBU},
            addressline={No. 1 International University Park Road}, 
            city={Shenzhen},
            postcode={518172}, 
            country={China}}

\fntext[equal]{Equal contribution.}
\cortext[cor]{Corresponding author.}

\begin{abstract}
Despite recent advances in Video Large Language Models (Vid-LLMs), Temporal Video Grounding (TVG) targeting precise localization of time segments of query events remains a grand challenge. Recent studies attempt to address this challenge by matching two separate tokens to all frame features and identifying the start and end frames of events based on the highest matching similarities. However, this approach relies heavily on the exact start and end timestamps, which are often ambiguous without capturing the semantic continuity and integrity of the event. To this end, we design E.M.Ground, a temporal grounding Vid-LLM that tackles the TVG challenge from a holistic and coherent event perception perspective. E.M.Ground introduces three novel designs: \textit{(i)} It introduces a special $<\!\!evt\!\!>$ token that aggregates information from all frames of query events with varying lengths, thereby preserving semantic continuity and integrity for accurate event matching. \textit{(ii)} It introduces Savitzky–Golay smoothing to suppress the noise and abrupt changes in token-to-frame similarities across adjacent timestamps, thereby improving prediction accuracy. \textit{(iii)} It aggregates multi-grained frame features to enhance the reliability of token-to-frame matching and improve temporal understanding, thereby compensating for the compression-induced information loss. \secondrevise{Extensive experiments on widely adopted TVG benchmarks, along with in-depth analyses, demonstrate that E.M.Ground achieves strong overall performance, with the most pronounced improvements on single-event grounding and boundary-integrity tasks.}
\end{abstract}


\end{frontmatter}

\input{1_introduction}
\input{2_related_work}
\input{3_method}
\input{4_experiment}

\bibliographystyle{elsarticle-num}
\bibliography{main}

\end{document}

%% file: 1_introduction.tex
\section{Introduction}

Video Large Language Models (Vid-LLMs)~\cite{xu2024pllava,lin2023video, li2024llama,wang2023internvid} have made great advances in general video understanding, such as general question answering~\cite{fu2024video,liu2024bench,liu2024triadic,chen2026super,liu2022dynamic} and video captioning~\cite{maaz2023video,ma2025retta,chen2026ask}. However, these models still face challenges in Temporal Video Grounding (TVG)~\cite{gao2017tall, liu2024bench}, a fine-grained task that requires precise timestamp localization of the query event~\cite{ren2024timechat, huang2024vtimellm}.
TVG is crucial for helping users quickly locate target segments and perform in-depth analyses without manually searching through the entire video~\cite{guo2024trace}.

\begin{figure}[t]
    \centering
    \vspace{-12mm}
    \includegraphics[width=0.78\linewidth]{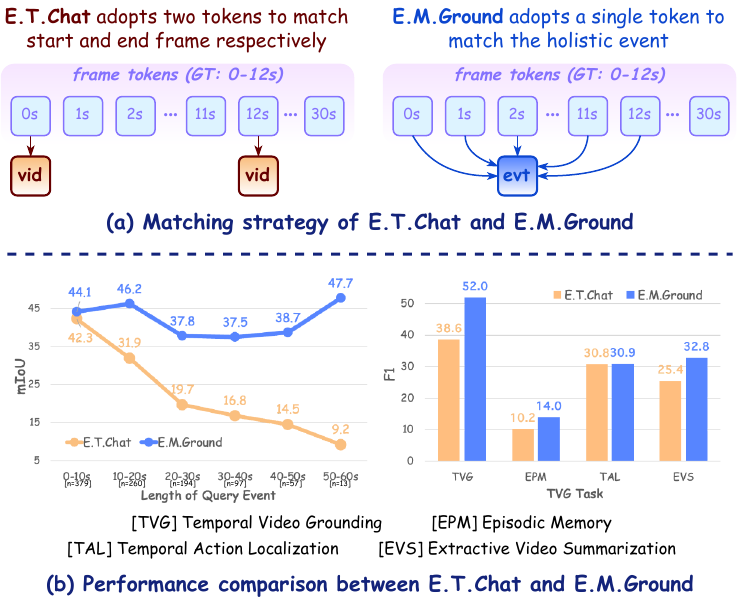}
    \vspace{-3mm}
    \caption{\revise{Comparison between E.T.Chat and our proposed E.M.Ground. (a) E.T.Chat~\cite{liu2024bench} uses two separate tokens to match the start and end frames, whereas E.M.Ground uses a unified \(<\!\!evt\!\!>\) token to match the complete event span. (b) Performance across different event-duration groups. The number in square brackets denotes the number of evaluated videos in each group, \textit{e.g.}, \(0\text{-}10\,\mathrm{s}\,[n=379]\), while the numerical annotations indicate the corresponding performance values. By preserving the semantic continuity and integrity of events, E.M.Ground achieves better and more stable performance across different event durations.}}
    \label{fig:motivation}
    \vspace{+4mm}
\end{figure}

\revise{Several representative TVG frameworks have been explored in the literature. Traditional methods formulate TVG as cross-modal moment localization, including proposal-based methods~\cite{gao2017tall}, temporal-map based methods~\cite{zhang2021multi}, and detection-based methods~\cite{lei2021detecting}. With the development of Vid-LLMs, recent studies attempt to solve TVG by directly generating textual timestamps~\cite{qu2024chatvtg,ren2024timechat,wang2024hawkeye,huang2024vtimellm} or by introducing explicit temporal representations, such as special time tokens and lightweight temporal decoders~\cite{huang2024lita,deng2024seq2time,guo2025vtg,guo2024trace,qian2024momentor}. These methods improve temporal awareness, but accurate timestamp prediction remains challenging due to the limited numerical reasoning ability of LLMs~\cite{dziri2023faith}, the computational cost of dense temporal representations, and the difficulty of modeling long videos~\cite{chen2025longvila,di2024grounded}.}

\revise{Different from methods that either generate textual timestamps directly or decode temporal boundaries from explicit time representations, E.T.Chat~\cite{liu2024bench} introduces a token-to-frame matching strategy. It predicts temporal boundaries by matching two special boundary tokens with frame-level visual features (refer to Fig.~\ref{fig:motivation}(a)). Specifically, E.T.Chat supervises two special tokens to match the start and end boundary frames. However, prior TVG studies have shown that temporal context and adjacent moment relations are important for localizing complete events~\cite{zhang2021multi,lei2021detecting}. In contrast, the intermediate frames that constitute the semantic body of the event are not explicitly modeled as a coherent event in such boundary-centric token matching. During inference, the predicted boundaries are further determined by token-to-frame similarity peaks, which can be sensitive to local fluctuations across adjacent frames. This issue becomes more pronounced for long videos, where aggressive visual compression may discard fine-grained temporal evidence~\cite{guo2025vtg,azad2025hierarq}. As shown in Fig.~\ref{fig:motivation}(b) and further analyzed in Fig.~\ref{fig:detail_tvg} and Fig.~\ref{fig:error_statistics}, E.T.Chat exhibits degraded performance on longer events and produces more boundary-related grounding errors. This observation highlights the necessity of developing a method that can perceive the holistic and coherent semantics of query events.}

To this end, we design E.M.Ground, a novel temporal grounding Vid-LLM that matches the query event from a holistic and coherent perception perspective. E.M.Ground comprises three matching designs. \textbf{\textit{First}}, it introduces a unified token $<\!\!evt\!\!>$ to represent the holistic query event, which is trained to match all frames within the ground-truth event span. This design preserves the semantic continuity and integrity of the query event. \textbf{\textit{Second}}, it introduces Savitzky-Golay smoothing~\cite{savitzky1964smoothing}, a technique widely used in signal processing~\cite{hamming1998digital,pandia2010motion, persson2003smoothing}, to suppress noise and abrupt changes in token-to-frame similarity across adjacent frames during inference. \textbf{\textit{Third}}, it aggregates multi-grained frame features, which compensates for the information loss brought by aggressive visual compression. Extensive experiments show that E.M.Ground effectively mitigates the limitations of the previous matching approach~\cite{liu2024bench}, yielding consistent performance gains across diverse TVG tasks and robust results on videos of varying lengths (refer to Fig.~\ref{fig:motivation}(b)).

Our contributions can be summarized in three major aspects: \revise{\textbf{\textit{First}}, we identify a fundamental limitation of the boundary-centric timestamp matching strategy~\cite{liu2024bench}: independently matching start and end frames fails to explicitly model the semantic continuity and integrity of the complete event. Accordingly, we introduce the principle of holistic event perception and matching. \textbf{\textit{Second}}, we instantiate this principle in E.M.Ground by aligning a unified \(<\!\!evt\!\!>\) token with all frames within the query event, aggregating multi-grained visual features, and applying Savitzky-Golay smoothing to suppress noisy response fluctuations.} \secondrevise{\textbf{\textit{Third}}, extensive experiments demonstrate that E.M.Ground effectively addresses the limitations of boundary-centric matching and achieves strong overall performance, particularly on single-event grounding and boundary-integrity tasks.}

%% file: 2_related_work.tex
\section{Related Works}

\subsection{Temporal Video Grounding}

\revise{Temporal Video Grounding (TVG) aims to localize the temporal segment in a video that corresponds to a natural-language query~\cite{gao2017tall}. Traditional TVG methods can be broadly categorized into three paradigms. \textit{(i)} Proposal-based methods generate candidate temporal segments and rank them according to their cross-modal relevance to the query~\cite{wang2022negative}. These methods benefit from explicit segment-level comparison, but their performance depends heavily on proposal quality and dense candidate generation. \textit{(ii)} Proposal-free methods directly regress the start and end timestamps of the target moment~\cite{zhang2020span}, which improves efficiency but still requires accurate boundary regression. \textit{(iii)} Detection-based methods formulate temporal grounding as moment detection and localize relevant segments with transformer-based detectors~\cite{lei2021detecting}. Although these traditional methods have advanced TVG, they are usually task-specific and do not fully exploit the instruction-following and open-ended reasoning ability of recent Vid-LLMs.}

\subsection{Video Large Language Models for TVG}

\revise{Recent studies leverage Video Large Language Models (Vid-LLMs) for TVG due to their strong video-language understanding ability~\cite{xu2024pllava,maaz2023video,wang2025time}. According to how temporal predictions are produced, existing Vid-LLM-based TVG methods can be divided into three categories.}

\subsubsection{\revise{Direct timestamp generation}}

\revise{Early Vid-LLM-based methods directly generate numerical timestamps as plain text~\cite{qu2024chatvtg,ren2024timechat,wang2024hawkeye,huang2024vtimellm}. This strategy is simple and compatible with the language-generation interface of LLMs. However, accurate timestamp generation remains challenging because LLMs are not inherently optimized for fine-grained numerical reasoning and temporal boundary regression~\cite{dziri2023faith,liu2024bench,guo2024trace}.}

\subsubsection{\revise{Temporal-token and decoder-based grounding}}

\revise{To improve temporal awareness, subsequent works introduce explicit temporal representations. Some methods represent timestamps with special numerical tokens~\cite{huang2024lita,deng2024seq2time,guo2025vtg}, while others decode timestamps from special temporal tokens with lightweight temporal decoders~\cite{guo2024trace,qian2024momentor}. These methods provide stronger temporal supervision than direct text generation, but they may require additional learnable modules, dense temporal representations, or carefully designed timestamp tokens, which increase training and inference complexity.}

\subsubsection{\revise{Token-to-frame matching}}

\revise{Different from the above paradigms, E.T.Chat~\cite{liu2024bench} introduces a token-to-frame matching strategy. It samples videos in an FPS-based manner and predicts the start and end timestamps by matching two special tokens with frame-level visual features. This design avoids direct numerical timestamp generation and achieves strong performance. However, it remains boundary-centric: the start and end tokens are supervised to match boundary frames, while the intermediate frames that constitute the semantic body of an event are not explicitly modeled as a coherent event. Moreover, the final prediction relies on token-to-frame similarity peaks, which can be sensitive to local fluctuations. In contrast, our E.M.Ground models the whole query event with a unified \(<\!\!evt\!\!>\) token and matches all frames within the target event span, thereby preserving event-level semantic continuity and integrity.}

\subsection{\revise{Structured Visual Alignment and Consistency Learning}}

\revise{Our work is also related to structured visual alignment and consistency learning. Recent studies show that explicit concepts, structural awareness, and consistency constraints can improve fine-grained visual understanding, as demonstrated by segmentation foundation models~\cite{carion2025sam3}, symmetrical awareness generation for pelvic image segmentation~\cite{song2025symmetrical}, and generative consistency learning for cerebrovascular segmentation~\cite{chen2022generative}. Although these works focus on different visual domains, they motivate us to align the \(<\!\!evt\!\!>\) token with the complete event span rather than isolated temporal boundaries.}

%% file: 3_method.tex
\section{Method}\label{sec:method}

\begin{figure*}[t]
\centering
    \vspace{-12mm}
    \includegraphics[width=\linewidth]{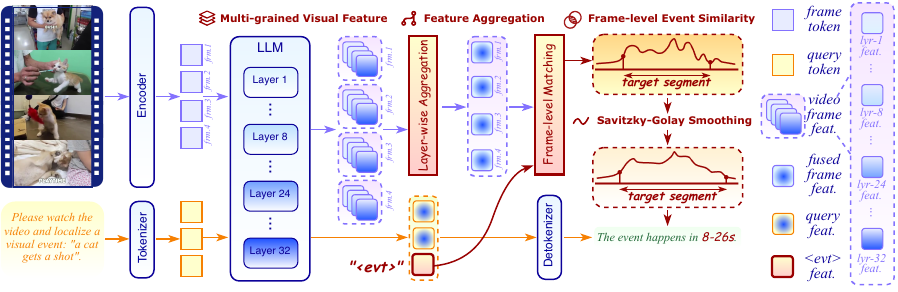}
    \vspace{-10mm}
    \caption{Overall architecture of our proposed E.M.Ground. Specifically, E.M.Ground perceives the query event from a holistic and coherent perspective. It introduces a special token $<\!\!evt\!\!>$ to aggregate all frames within the ground-truth time spans, leverages multi-grained visual features to compensate for information loss, and refines the predictions with smoothing operations. Based on these designs, E.M.Ground effectively captures the semantic continuity and integrity of the query event.}
    \vspace{+4mm}
\label{fig:model}
\end{figure*}

The proposed E.M.Ground is illustrated in Fig.~\ref{fig:model}. In this section, we elaborate on three major designs as follows: \textit{(i)} leveraging multi-grained visual features for enhanced representation, \textit{(ii)} introducing holistic and coherent event matching for improved grounding accuracy, and \textit{(iii)} adopting Savitzky-Golay smoothing to refine predicted segments.

\subsection{Multi-grained Visual Feature Extraction}\label{ssec:multi-grain}
Given an input $T$-frame video $\mathbf{V} = \{\mathbf{V}_{1}, \mathbf{V}_{2}, \dots, \mathbf{V}_{T}\} \in \mathbb{R}^{T \times H \times W \times 3}$, where $H$ and $W$ denote the height and width of each frame, we extract a sequence of frame tokens $\mathbf{v} = \{\mathbf{v}_{1}, \mathbf{v}_{2}, \dots, \mathbf{v}_{T}\}$, where $\mathbf{v}_{t} \in \mathbb{R}^{1 \times D}$ represents the visual token of frame $t$. We describe the details of the visual encoder in Sec.~\ref{ssec:training}. Similar to prior Vid-LLMs~\cite{ren2024timechat,guo2024trace,liu2024bench,azad2025hierarq}, we compress the visual tokens to meet the computational constraints of video understanding tasks. To mitigate the resulting information loss, we incorporate multi-grained visual features to enrich the representation~\cite{lin2025multi,yao2024dense}. As an example, consider the frame token $\mathbf{v}_{t}$ corresponding to frame $t$:
\begin{equation}
\begin{aligned}
    \mathbf{f}_{vt} = \{\mathbf{f}^{1}_{vt}, \mathbf{f}^{2}_{vt}, \dots, \mathbf{f}^{L}_{vt}\},\ \ \ \ \ 
    \mathbf{f}^{l}_{vt} = M^{l}(\mathbf{v}_t),
\end{aligned}
\end{equation}
where $M$ denotes the LLM, $M^{l}$ represents the features from the $l$-th layer ($0 \leq l \leq L$),
$\mathbf{f}^{l}_{vt}$ represents the frame features from the $l$-th layer, and $\mathbf{f}_{vt} \in \mathbb{R}^{L \times D}$ denotes the multi-grained features of frame $t$. We then aggregate these features across layers as follows:
\begin{equation}
\begin{aligned}
    \overline{\mathbf{f}_v} = \{\overline{\mathbf{f}_{v1}}, \overline{\mathbf{f}_{v2}}, \dots, \overline{\mathbf{f}_{vT}}\},\ \ \ \ \ 
    \overline{\mathbf{f}_{vt}} = \frac{1}{L} \sum_{l=1}^{L} M^{l}(\mathbf{v}_t),
\end{aligned}
\end{equation}
where $\overline{\mathbf{f}_{vt}} \in \mathbb{R}^{1 \times D}$ is the aggregated feature of frame $t$.

\subsection{Holistic and Coherent Event Matching}
\subsubsection{Overview} Since the previous timestamp matching strategy proposed in E.T.Chat~\cite{liu2024bench} is suboptimal, we match the query event from a holistic and coherent perception perspective instead. Specifically, we define a special token $<\!\!evt\!\!>$ to serve as the matching signal. For example, in the TVG task, the expected output format is: ``The event happens in $<\!\!evt\!\!>$''. The special token $<\!\!evt\!\!>$ is used to match all frames and identify candidate segments based on similarity score across the entire video. Specifically, we utilize the $<\!\!evt\!\!>$ token's features from the second-to-last layer for event matching, which retains a broader feature range~\cite{liu2024bench,he2024multi}.

\subsubsection{Details} First, the extracted $<\!\!evt\!\!>$ feature, denoted as $\mathbf{f}^{L-1}_{\text{evt}}$, and the aggregated visual features $\overline{\mathbf{f}_{v}}$ are each passed through separate projectors for modality alignment prior to similarity computation. We adopt two MLPs $E_{evt}$ and $E_{v}$ serving as the projectors:
\begin{equation}
\begin{aligned}
    \mathbf{m}_{evt} = E_{evt}(\mathbf{f}^{L-1}_{\text{evt}}),\ \ \ \ \ 
    \mathbf{m}_v = E_v(\overline{\mathbf{f}_v}),
\end{aligned}
\end{equation}
where $\mathbf{m}_{evt} \in \mathbb{R}^{1 \times D}$ represents the $<\!\!evt\!\!>$ feature for matching, and $\mathbf{m}_{v} \in \mathbb{R}^{T \times D}$ represents the frame-level visual features in the same alignment space. Next, we compute the cosine similarity between the $\mathbf{m}_{evt}$ and $\mathbf{m}_v$:
\begin{equation}
    sim = \frac{\mathbf{m}_{evt} \cdot \mathbf{m}_v}{||\mathbf{m}_{evt}||_2 \cdot ||\mathbf{m}_v||_2}.
\end{equation}
Here, $sim = \{\textbf{s}_1,\textbf{s}_2,\dots,\textbf{s}_T\} \in \mathbb{R}^{1 \times T}$ denotes the similarity between the $<\!\!evt\!\!>$ token and each frame of the input video, and $\textbf{s}_t$ denotes the similarity score of frame $t$. Subsequently, we predict the target segments based on $sim$:
\begin{equation}
\begin{aligned}
    seg = \{\text{(start,\ end)}|\textbf{s}_t>\sigma\ \forall t \in \text{[start,\ end]},\ 
    \text{and}\ \textbf{s}_{\text{start}-1} \leq \sigma, \textbf{s}_{\text{end}+1} \leq \sigma\},
\end{aligned}    
\end{equation}
where ``start'' and ``end'' represent the boundary timestamps of predictions. Finally, the $<\!\!evt\!\!>$ in the original output is replaced by the boundary of predicted segments. For instance, if the target segment is ``(8, 26)'', the final output would be ``The event happens in 8 – 26s''.

\subsubsection{Ground-truth labels} We adopt a smoothed binary label to construct the ground-truth time slot representation $\mathbf{y}_t$, promoting numerical continuity. Specifically, all timestamps within the target event are labeled as $1$, and a smoothing strategy is applied to those outside the boundaries to ensure continuity and reduce abrupt transitions:
\begin{equation}\label{eq:gt_label}
    \mathbf{y}_t =
    \begin{cases}
    1, & \text{if } s \leq t \leq e \\
    \frac{1}{\alpha^{\min(|t - s|, |t - e|)}}, & \text{else if}\ \revise{\min(|t - s|, |t - e|) \leq d} \\
    0, & \text{otherwise},
    \end{cases}
\end{equation}
where $s$ and $e$ represent the start and end timestamps of the ground-truth time slot, respectively. For the VHD task, $s = e$. \revise{Following E.T.Chat, we set \(\alpha=2\) and apply label smoothing only within a small boundary neighborhood \(d=3\), \textit{i.e.}, \(\min(|t-s|,|t-e|)\leq 3\). Since videos are sampled at \(1\) FPS for frame-level matching, this corresponds to a three-frame temporal neighborhood around the event boundary. With \(\alpha=2\), the smoothed labels decay rapidly from \(0.5\) to \(0.25\) and \(0.125\), so frames farther away from the event receive negligible supervision.}  For example, for a 12-second video with the ground-truth time slot from 5 to 10 seconds, the target representation is defined as: $\mathbf{y}_t = [0, 0.125, 0.25, 0.5, 1, 1, 1, 1, 1, 1, 0.5, 0.25]$.

\subsubsection{Loss function} We incorporate the standard Negative Log-Likelihood over generated tokens widely employed in multi-modal LLMs~\cite{liu2023visual,lin2023video,maaz2023video,liu2024bench}:
\begin{equation}
    p(\mathbf{X}_a \mid \mathbf{X}_v, \mathbf{X}_q) 
    = \prod_{i=1}^{L} p_\theta \big(x_i \mid \mathbf{X}_v, \mathbf{X}_q, \mathbf{X}_{a,<i} \big),
\end{equation}
where $\mathbf{X}_v$ and $\mathbf{X}_q$ denote the visual and textual inputs, respectively, $\mathbf{X}_a$ represents the answer tokens, $\mathbf{X}_{a,<i}$ indicates all previously generated answer tokens, and $\theta$ is the trainable model parameters. The corresponding Negative Log-Likelihood Loss is:
\begin{equation}
    \mathcal{L}_{\text{NLL}} = - \sum_{i=1}^{L} \log p_\theta \big(x_i \mid \mathbf{X}_v, \mathbf{X}_q, \mathbf{X}_{a,<i} \big).
\end{equation}
Additionally, we also introduce an auxiliary loss specifically designed for the event matching task.
\revise{Before calculating the matching loss, we normalize the cosine similarity scores over all frames using softmax:}

\begin{equation}
    \revise{p_t = \frac{\exp(s_t)}{\sum_{j=1}^{T}\exp(s_j)},}
\end{equation}
\revise{where \(p_t>0\) and \(\sum_{t=1}^{T}p_t=1\). \secondrevise{No additional temperature scaling is used, which is equivalent to setting \(\tau=1\). For a fixed cosine-similarity sequence, temperature scaling changes the sharpness of the softmax distribution but does not change the relative ordering of frames. In our method, \(p_t\) is used only as a video-wise normalized variable for the matching loss rather than as a calibrated binary foreground probability. Through the shared softmax normalization, the soft-label cross-entropy loss increases the relative scores of frames within the target event while suppressing those of non-event frames. Since the supervision covers the complete event span rather than a single peak frame, we retain \(\tau=1\) without introducing an additional temperature hyperparameter. The probability \(p_t\) is used exclusively for calculating the matching loss and is not used for temporal boundary prediction.} The matching loss is then formulated as a soft-label cross-entropy loss:}

\begin{equation}
    \revise{\mathcal{L}_{\mathrm{matching}}
= -\frac{1}{T}\sum_{t=1}^{T}y_t\log(p_t),}
\end{equation}
\revise{where \(y_t\) denotes the smoothed event label defined in Eq.~\ref{eq:gt_label}. The logarithm is applied to the normalized probability \(p_t\). Although \(y_t=0\) for non-event frames, their probabilities are suppressed through competition with event frames in the softmax normalization.} \secondrevise{Following E.T.Chat~\cite{liu2024bench}, the overall training objective is formulated as: $\mathcal{L} = \mathcal{L}_{\mathrm{NLL}} + \mathcal{L}_{\mathrm{matching}}.$
The two objectives provide complementary supervision: \(\mathcal{L}_{\mathrm{NLL}}\) preserves language generation and instruction following, whereas \(\mathcal{L}_{\mathrm{matching}}\) introduces frame-level temporal supervision for event grounding. Their gradient pathways are partially separated. The language-modeling head is supervised by the NLL loss, while the event and frame projectors are supervised by the matching loss. The two objectives interact only through the shared representations, thereby limiting direct cross-objective interference in the objective-specific modules. Following E.T.Chat, we directly optimize the equally weighted objective without additional gradient-balancing or gradient-surgery operations.}

\subsection{Savitzky-Golay Smoothing}\label{ssec:sg_smooth}
Since an event typically contains rich information, including subjects, actions, and environments, using a single $<\!\!evt\!\!>$ token to represent and match the entire event remains a significant challenge. This leads to several limitations: \textit{(i)} some event segments may exhibit relatively low similarity with the $<\!\!evt\!\!>$ token; \textit{(ii)} the start and end phases of the event may show reduced similarity; and \textit{(iii)} the similarity transition between adjacent frames can be abrupt. Therefore, we incorporate Savitzky-Golay smoothing~\cite{savitzky1964smoothing} to further enhance grounding accuracy. This technique offers several advantages: \textit{(i)} it effectively filters out noise and abrupt variations~\cite{hamming1998digital}; \textit{(ii)} it preserves peak shapes~\cite{pandia2010motion}, which is crucial for detecting short-duration events and performing well on the VHD task; and \textit{(iii)} it demonstrates strong performance in time-domain signal processing~\cite{persson2003smoothing}. \revise{The expression of Savitzky-Golay smoothing for each similarity score \(s_t\) is:
\begin{equation}
    \tilde{s}_t = \sum_{i=-k}^{k} c_i^{(k,p)} s_{t+i},
\end{equation}
where \(k\) denotes the window size, \(p\) denotes the polynomial order, and \(c_i^{(k,p)}\) denotes the convolution coefficient determined by \(k\) and \(p\). These coefficients are computed by performing a least-squares polynomial fit of order \(p\) over a local window of \(2k+1\) consecutive points centered at time \(t\). The order \(p\) controls the local shape-preserving ability of the filter, while the window size \(k\) controls the smoothing range. In all experiments, we set \(p=2\) and \(k=5\) by default and analyze the influence of \(k\) and \(p\) in Sec.~\ref{ssec:hpt}.} The filter preserves peaks and edges while reducing high-frequency noise. \revise{The $sim$ after smoothing is:
\begin{equation}
    \widetilde{sim} = S_{\mathrm{S\mbox{-}G}}(\mathrm{sim}; k,p),
\end{equation}
where \(\widetilde{sim}=\{\tilde{s}_1,\tilde{s}_2,\ldots,\tilde{s}_T\}\), and
 \(S_{\mathrm{S\mbox{-}G}}\) represents Savitzky-Golay smoothing.} 
\secondrevise{During inference, Savitzky--Golay smoothing is applied directly to the cosine-similarity sequence \(sim=\{s_t\}_{t=1}^{T}\), rather than to the softmax probabilities \(p_t\). The refined target segments are therefore obtained by applying the threshold \(\sigma\) to the smoothed cosine similarities \(\tilde{s}_t\):}
\begin{equation}
\begin{aligned}
    \widetilde{seg} = \{\text{(start,\ end)}|\tilde{\textbf{s}}_t>\sigma\ \forall t \in \text{[start,\ end]},\ 
    \text{and}\ \tilde{\textbf{s}}_{\text{start}-1} \leq \sigma, \tilde{\textbf{s}}_{\text{end}+1} \leq \sigma\}.
\end{aligned}
\end{equation}
\secondrevise{Thus, \(\sigma\) serves as a near-zero decision boundary in the cosine-similarity space and is independent of the training-time softmax normalization. As analyzed in Sec.~\ref{ssec:hpt}, E.M.Ground maintains stable performance across a range of near-zero values of \(\sigma\).}

\subsection{Model Components and Training Details}\label{ssec:training}
\noindent

\subsubsection{Model components} We adopt the pre-trained ViT-G/14~\cite{dosovitskiy2020image} from EVA-CLIP~\cite{fang2023eva} as the vision encoder. For visual compression, we engage Q-Former~\cite{dai2023instructblip,li2023blip} to reduce the memory occupancy. The Phi-3-Mini-3.8B~\cite{abdin2024phi} is chosen as our LLM backbone. Details are provided in the supplementary materials.

\subsubsection{Training details} We follow most of the training strategy and data utilization from LLaMA-VID~\cite{maaz2023video} and E.T.Chat~\cite{liu2024bench}. In stage-1, we focus on video-text alignment and only unfreeze the aggregator and projector. Consequently, large-scale coarse-aligned data (\textit{i.e.}, WebVid~\cite{bain2021frozen} and LCS-558K~\cite{liu2023visual}) is adopted in this stage. The stage-2 aims to improve the capability of instruction following; thus, the Q-Former and LLM are also unfrozen. We leverage ActivityNet~\cite{caba2015activitynet} and LLaVA-1.5-Instruct~\cite{liu2023visual} as the instruction data. We unfreeze the projector for event matching in stage-3, and conduct a two-epoch training with related data from E.T.Instruct-164K~\cite {liu2024bench} for grounding tasks.  We train the model with mixed precision (FP16) on a compute node with 8 × NVIDIA A800 GPUs.

\subsection{Advantages of E.M.Ground}
Compared with previous temporal grounding Vid-LLMs, E.M.Ground provides several key advantages: \textit{(i)} it perceives the event span using a single token, $<\!\!evt\!\!>$, rather than directly generating numerical timestamps or relying on a long token sequence; \textit{(ii)} it predicts temporal boundaries based on token-to-frame similarity, without introducing additional learnable parameters; \textit{(iii)} it captures the holistic and coherent semantics of the query event, instead of over-relying on the start and end boundaries; \textit{(iv)} it leverages multi-grained visual features to compensate for information loss introduced by visual compression; and \textit{(v)} it refines predictions using smoothing operations to suppress noise and abrupt changes across adjacent frames.

\input{tab_tvg.tex}

%% file: tab_tvg.tex
\begin{table}[t]
    \renewcommand\arraystretch{0.9}
    \centering
    \vspace{-12mm}
    \caption{Zero-shot performance on the Temporal Video Grounding task is evaluated using the Charades-STA~\cite{gao2017tall} (\textbf{Up}) and relevant E.T.Bench~\cite{liu2024bench} subtasks (\textbf{Down}). The best results are highlighted in \textbf{bold} font.}
    \vspace{+4mm}
    
        \setlength{\tabcolsep}{18pt}
        \scalebox{0.78}{
        \begin{tabular}{l|ccc|c}
        \toprule[1pt]
        \multicolumn{5}{c}{\textit{\textbf{Charades-STA}}}\\\hline
        Methods& R@0.3& R@0.5& R@0.7& mIoU\\\hline\hline
        NumPro (7B)~\cite{wu2024number}   & 27.2& 10.3& 2.9& 18.9\\
        Momentor (7B)~\cite{qian2024momentor} & 42.6& 26.6& 11.6& 28.5\\
        Seq2Time (7B)~\cite{deng2024seq2time} & -& 31.2& 13.7& -\\
        HawkEye (7B)~\cite{wang2024hawkeye}  & 50.6& 31.4& 14.5& 33.7\\
        TimeChat (7B)~\cite{ren2024timechat} & -& 32.2& 13.4& -\\
        ChatVTG (7B)~\cite{qu2024chatvtg}  & 52.7& 33.0& 15.9& 34.9\\
        VTG-LLM (7B)~\cite{guo2025vtg}  & -& 33.8& 15.7& -\\
        VTimeLLM (13B)~\cite{huang2024vtimellm} & 55.3& 34.3& 14.7& 34.6\\
        TRACE (7B)~\cite{guo2024trace}    & -& 40.3& \textbf{19.4}& -\\
        \cellcolor{mycolor}\textbf{Ours (3.8B)}   & \cellcolor{mycolor}\textbf{67.1}& \cellcolor{mycolor}\textbf{44.8}& \cellcolor{mycolor}\textbf{19.4}& \cellcolor{mycolor}\textbf{43.3}\\
        \bottomrule[1pt]
        \end{tabular}
        }

        \vspace{+2mm}

        \setlength{\tabcolsep}{20pt}
        \scalebox{0.78}{
        \begin{tabular}{l|c|c|c|c}
        \toprule[1pt]
        \multicolumn{5}{c}{\textit{\textbf{E.T.Bench-Grounding}}}\\\hline
        \multirow{2}*{Methods}& TVG& EPM& TAL& EVS\\\cline{2-5}
        ~& F1& F1& F1& F1\\\hline\hline
        LLaMA-VID (7B)~\cite{li2024llama}   & 5.5& 1.2& 8.0& 1.4\\
        PLLaVA (7B)~\cite{xu2024pllava}      & 6.9& 1.1& 5.7& 0.3\\
        Video-LLaVA (7B)~\cite{lin2023video} & 7.0& 1.9& 15.0& 0.3\\
        Video-ChatGPT(7B)~\cite{maaz2023video} & 7.0& 1.3& 15.1& 8.4\\
        VTimeLLM (7B)~\cite{huang2024vtimellm}    & 7.6& 1.9& 18.2& 15.9\\
        VTG-LLM (7B)~\cite{guo2025vtg}     & 15.9& 3.7& 14.4& 26.8\\
        LITA (7B)~\cite{huang2024lita}        & 26.2& 3.9& 10.1& 29.1\\
        E.T.Chat (3.8B)~\cite{liu2024bench}  & 38.6& 10.2& 30.8& 25.4\\
        \cellcolor{mycolor}\textbf{Ours (3.8B)}      & \cellcolor{mycolor}\textbf{52.0}& \cellcolor{mycolor}\textbf{14.0}& \cellcolor{mycolor}\textbf{30.9}& \cellcolor{mycolor}\textbf{32.8}\\
        \bottomrule[1pt]
        \end{tabular}
        }
    
    
    \label{tab:tvg}
\end{table}

%% file: 4_experiment.tex
\section{Experiment}\label{sec:experiment}

\subsection{Tasks, Datasets, and Evaluation Metrics}
\subsubsection{Temporal Video Grounding (TVG)} Temporal Video Grounding (TVG) requires determining the start and end timestamps of the query event within a given video~\cite{gao2017tall}. We adopt two TVG datasets, Charades-STA~\cite{gao2017tall} and E.T.Bench~\cite{liu2024bench} related subtasks (\textit{i.e.}, TVG, EPM, TAL, and EVS). \revise{Some subtasks in E.T.Bench~\cite{liu2024bench} contain multiple target segments, which require post-processing for LLM-generated responses. For a fair comparison, we strictly follow the same processing as E.T.Chat~\cite{liu2024bench}. During training, the model learns to generate one \(<\!\!evt\!\!>\) token for each identified event and aligns each token with its corresponding temporal span. Therefore, the number of generated \(<\!\!evt\!\!>\) tokens represents the number of events predicted by E.M.Ground. For each generated token, E.M.Ground independently computes a query-conditioned frame-similarity sequence and extracts its corresponding span using the thresholding strategy. Consequently, nearby events are represented and localized by different \(<\!\!evt\!\!>\) tokens.} More details about E.T.Bench are described in the supplementary materials. For Charades-STA, we follow the evaluation settings outlined in TimeChat~\cite{ren2024timechat} and report the mIoU and recall at IoU thresholds of 0.3, 0.5, and 0.7. For E.T.Bench, we report the F1-score for each subtask.

\subsubsection{Dense Video Captioning (DVC)} Dense Video Captioning (DVC) involves generating descriptive captions for multiple events within a video while simultaneously localizing their temporal boundaries~\cite{krishna2017dense}. We adopt two DVC datasets, YouCook2~\cite{zhou2018towards} and E.T.Bench~\cite{liu2024bench} related subtasks (\textit{i.e.}, DVC and SLC). For YouCook2, we follow the same settings as TimeChat and report performance under three metrics: \textit{(i)} CIDEr~\cite{vedantam2015cider} evaluates caption quality; \textit{(ii)} SODA\_c~\cite{fujita2020soda} for story-level assessment; and \textit{(iii)} F1-score measures the ability of identifying timestamps. For E.T.Bench, we report both the F1-score to evaluate temporal boundary accuracy and the Sim-score to assess caption correctness. We also leverage two subtasks from E.T.Bench to assess the captioning capabilities of E.M.Ground. \textbf{[DVC]} Dense Video Captioning: Generate comprehensive descriptions for all events occurring in the video. This represents a general setting, aiming to cover as many events as possible. \textbf{[SLC]} Step Localization and Captioning: Identify and describe only the key steps in instructional videos. Compared to [DVC], the segments in this setting are shorter, disjoint, and require more precise and focused descriptions. The prompts used for evaluation are provided in the supplementary materials.

\subsubsection{Video Highlight Detection} Video Highlight Detection (VHD) focuses on identifying customized moments and highlights within videos based on given text queries~\cite{lei2021detecting}. We adopt the QVHighlights dataset~\cite{lei2021detecting} and report the metrics used in TimeChat~\cite{ren2024timechat}, including mean Average Precision (mAP) at various IoU thresholds and HIT@1, which measures the hit rate of the top-ranked clip.

\subsection{Comparison with SoTA Methods on TVG Task}
In Tab.~\ref{tab:tvg}, we present the performance of our E.M.Ground in comparison with existing temporal grounding Vid-LLMs~\cite{wu2024number,qian2024momentor,deng2024seq2time,wang2024hawkeye,ren2024timechat,qu2024chatvtg,guo2025vtg,huang2024vtimellm,guo2024trace,li2024llama,xu2024pllava,maaz2023video,liu2024bench} across two representative TVG benchmarks, E.T.Bench~\cite{liu2024bench} and Charades-STA~\cite{gao2017tall}. 
\secondrevise{All evaluations follow the zero-shot protocols of TimeChat~\cite{ren2024timechat} and E.T.Chat~\cite{liu2024bench}, meaning that no training samples from the evaluation datasets are accessed during training. In particular, the video-ID intersection between E.T.Instruct-164K and the E.T.Bench test set is zero. For model-scale fairness, E.M.Ground uses a 3.8B LLM, tied with E.T.Chat for the smallest model scale among all compared methods, while the remaining methods use 7B or 13B models. Therefore, our performance advantage cannot be attributed to a larger model scale.} The results demonstrate that E.M.Ground consistently achieves new state-of-the-art performance across all metrics. Specifically, E.M.Ground outperforms TRACE~\cite{guo2024trace} by 4.5\% on R@0.5 in Charades-STA~\cite{gao2017tall}, and surpasses E.T.Chat~\cite{liu2024bench} by 13.4\% and 7.4\% on the TVG and EVS subtasks of E.T.Bench~\cite{liu2024bench}, respectively. It is noteworthy that E.M.Ground utilizes a relatively small language model (Phi-3 Mini-3.8B~\cite{abdin2024phi}), yet outperforms a range of Vid-LLMs with larger model sizes~\cite{huang2024vtimellm,ren2024timechat,guo2025vtg}. Moreover, E.M.Ground demonstrates strong capabilities in handling scenarios involving multiple target segments, \textit{i.e.}, the TAL and EVS subtasks in E.T.Bench.

\begin{figure*}[t]
\centering
    \vspace{-12mm}
    \includegraphics[width=\linewidth]{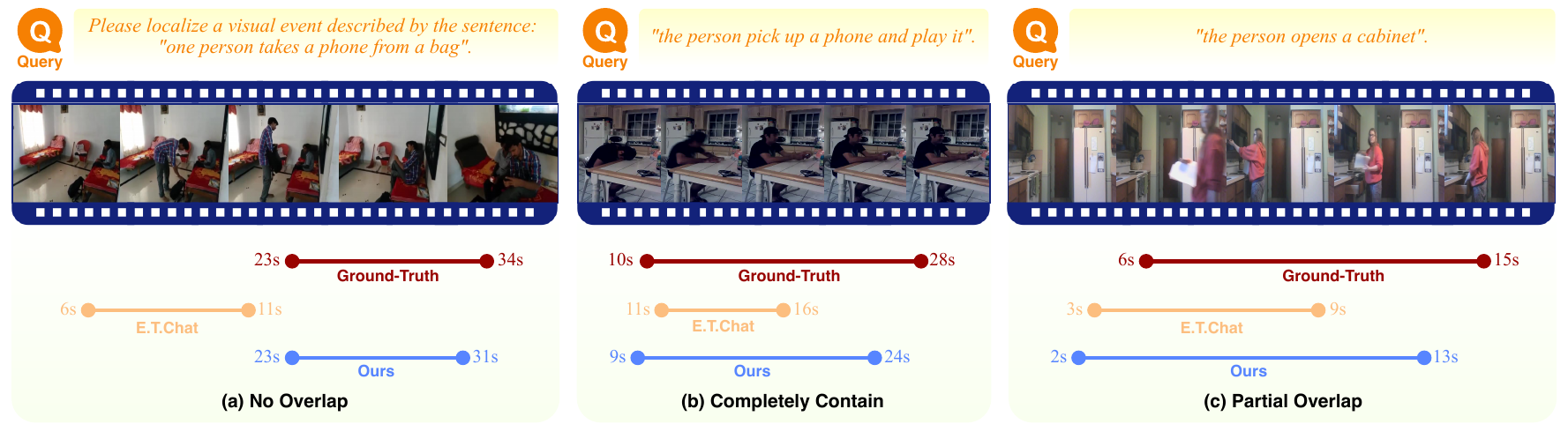}
    \vspace{-8mm}
    \caption{\revise{Qualitative comparison between E.T.Chat~\cite{liu2024bench} and E.M.Ground. Each example shows the query, video timeline, ground-truth segment (red bar), prediction of E.T.Chat (yellow bar), and the prediction of E.M.Ground (blue bar). The start and end timestamps beside each bar indicate the temporal boundaries of the corresponding segment. We present three representative failure modes of E.T.Chat, which are detailed in Sec.~\ref{ssec:error_discussion}. (a) \textbf{No Overlap}: E.T.Chat focuses on visually related objects, such as the person and the bag, but misses the queried action of taking a phone from the bag, resulting in a predicted segment that does not overlap with the ground-truth event. (b) \textbf{Completely Contained}: E.T.Chat only covers the middle part of the event and omits the beginning and ending phases, leading to an incomplete prediction fully contained within the ground-truth segment. (c) \textbf{Partial Overlap}: E.T.Chat partially overlaps with the ground-truth segment but includes irrelevant temporal regions and misses part of the target event. In contrast, our proposed E.M.Ground effectively mitigates all of these errors, providing more accurate temporal grounding.}}
    \vspace{+4mm}
\label{fig:detail_tvg}
\end{figure*}

\subsection{Discussion on Errors in TVG}~\label{ssec:error_discussion}
\begin{figure}[t]
\centering
    \vspace{-12mm}
    \includegraphics[width=0.6\linewidth]{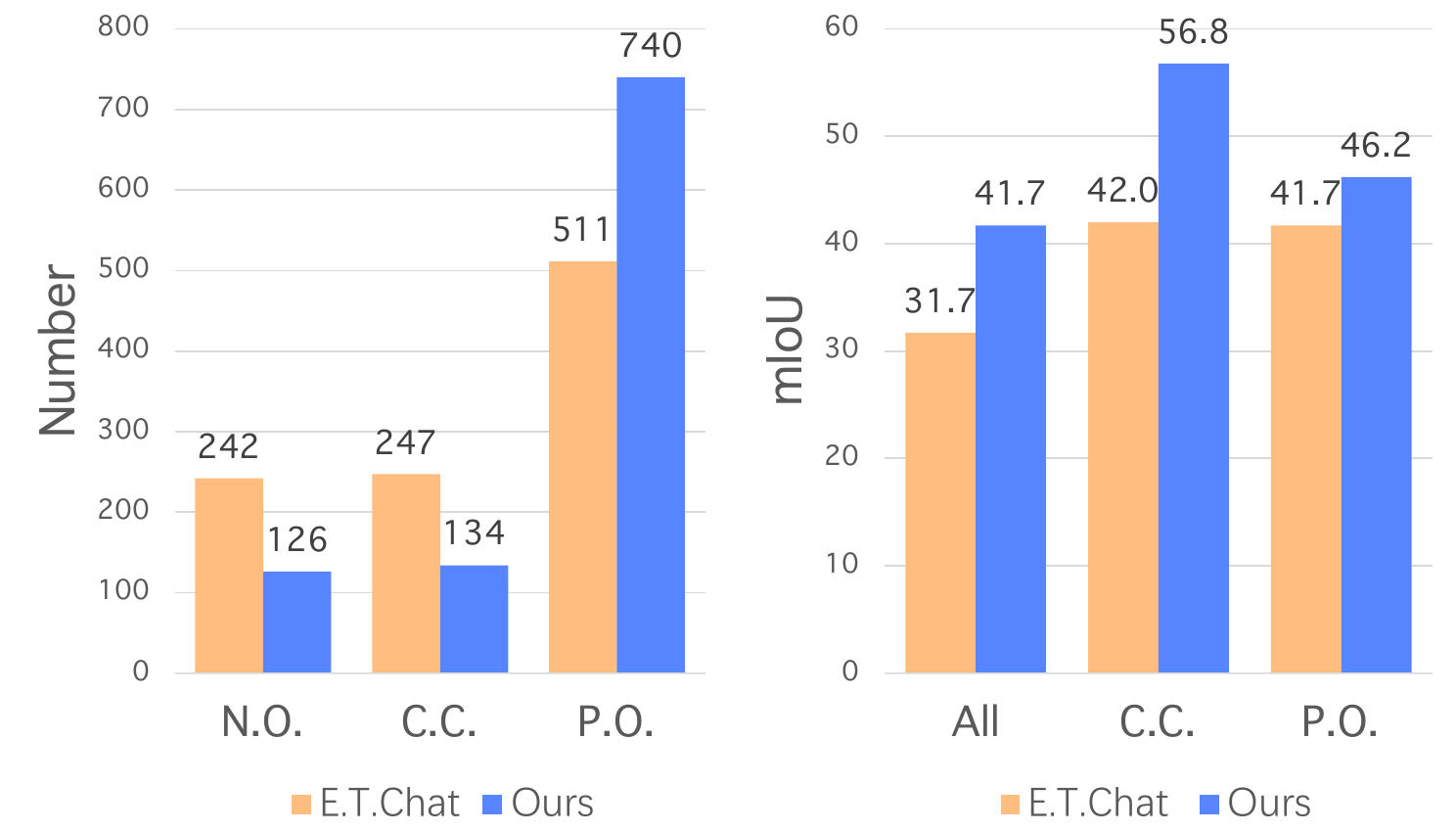}
    \vspace{-3mm}
    \caption{Error analysis of Temporal Video Grounding. \textbf{Left}: Number of errors for each type made by E.T.Chat~\cite{liu2024bench} and E.M.Ground. \textbf{Right}: mIoU corresponding to each error type. \textit{N.O.} denotes predictions that have no overlap with the ground-truths; \textit{C.C.} denotes cases where the ground-truths are completely contained within the predictions; \textit{P.O.} denotes predictions partially overlap with the ground-truths.}
    \vspace{+4mm}
\label{fig:error_statistics}
\end{figure}

In Fig.~\ref{fig:detail_tvg}, we analyze and categorize the errors in TVG tasks into three types. Detailed statistics from the TVG subtask of E.T.Bench are presented in Fig.~\ref{fig:error_statistics}. The first error type in Fig.~\ref{fig:detail_tvg}(a) refers to predictions that have no overlap with the ground-truth segments (\textit{i.e.}, mIoU = 0). We observe that 242 out of 1,000 questions predicted by E.T.Chat~\cite{liu2024bench} fall into this category, indicating a significant failure mode. This issue arises when timestamp matching is performed without considering the integrity of the target event. For example, the predictions from E.T.Chat in Fig.~\ref{fig:detail_tvg}(a) focus solely on the objects \textit{person} and \textit{bag}, while neglecting the happening of the action \textit{take}. Since the start and end timestamps are predicted independently, the model may select isolated timestamps corresponding to visually similar scenes, but which are not temporally connected. As a result, the predicted segments fail to align with the actual events described in the query. The second type involves predictions that are completely contained within the ground-truth segments, as illustrated in Fig.~\ref{fig:detail_tvg}(b). This occurs in 247 out of 1,000 samples, suggesting that E.T.Chat tends to focus on the middle stage of events while neglecting their beginning and ending phases, thereby compromising the integrity of the entire event. The predictions in Fig.~\ref{fig:detail_tvg}(b) cover only the first half of the ground-truth segments, failing to capture the complete event. The third error type in Fig.~\ref{fig:detail_tvg}(c) refers to predictions that are partially overlapped with the ground-truth segments but lack accurate alignment. These cases reflect insufficient temporal precision and incomplete event coverage. Compared to timestamp matching adopted by E.T.Chat, our proposed E.M.Ground significantly reduces the number of errors in the first two categories, while consistently improving the mIoU across all conditions, as illustrated in Fig.~\ref{fig:error_statistics}. \secondrevise{We further conduct a sample-level paired comparison between E.T.Chat and E.M.Ground on the same 1,000 TVG samples. Among the 242 no-overlap errors produced by E.T.Chat, E.M.Ground resolves 130, while introducing 14 new cases among the remaining 758 samples, resulting in 126 no-overlap errors in total. An exact two-sided McNemar's test, which tests whether resolving an existing error and introducing a new error are equally likely, confirms that E.M.Ground resolves significantly more no-overlap errors than it introduces (\(p=9.87\times10^{-25}\)). For the completely-contained category, 148 of the 247 errors produced by E.T.Chat no longer fall into this category, while 35 new cases are introduced among the remaining 753 samples, resulting in 134 cases in total. The corresponding McNemar's test also confirms a statistically significant reduction (\(p=9.74\times10^{-18}\)).} This indicates that our event matching approach effectively captures the semantic continuity and integrity of events, thereby enhancing the model's temporal grounding capabilities as intended.

\input{tab_intermediate_distractor}

\subsection{\secondrevise{Event-Length Performance and Causal Analysis of Event Coherence}}

In Fig.~\ref{fig:motivation}(b), we compare the proposed E.M.Ground with E.T.Chat~\cite{liu2024bench} on query events of varying lengths. E.M.Ground consistently outperforms E.T.Chat across all event durations. In contrast, E.T.Chat exhibits decreased performance on longer events, highlighting the limitation of its timestamp matching strategy: by focusing solely on start and end boundaries, it overlooks the richer information contained within the main body of longer events. E.M.Ground, on the other hand, maintains relatively stable performance across events of different lengths, further demonstrating the superiority of our proposed approach. \secondrevise{To directly examine whether the performance difference is related to coherent intermediate-frame modeling, we conduct a controlled intervention on the TVG subtask of E.T.Bench. Given an original ground-truth event \(\{v_s,\ldots,v_{e-1},v_e\}\), we remove its ending frame \(v_e\) and relocate it to a randomly selected later position outside the event, with non-event frames separating it from the remaining event sequence. All other frames retain their original content and temporal order. After relocation, \(v_{e-1}\) becomes the new correct ending frame, while the relocated \(v_e\) retains the visual appearance of the original boundary but is no longer temporally connected to the event. This intervention therefore separates the isolated boundary-frame cue from the coherent event sequence. As shown in Tab.~\ref{tab:intermediate_distractor}, the F1 score of E.T.Chat decreases from 38.6 to 30.7 after the intervention. It still selects the relocated original ending frame in 62.4\% of the samples and selects the new coherent endpoint in only 22.6\%. In contrast, E.M.Ground decreases from 52.0 to 49.9, selects the new endpoint in 68.7\% of the samples, and follows the relocated frame in only 11.8\%. These results show that E.T.Chat is strongly affected by the isolated appearance of the original boundary frame, whereas E.M.Ground determines the event endpoint primarily from the continuity and integrity of the complete event. Together with Fig.~\ref{fig:motivation}(b), this intervention demonstrates that insufficient modeling of intermediate-frame coherence contributes to the degraded performance of E.T.Chat on longer events.}

\input{tab_ablation.tex}

\subsection{Ablation Study}
We conduct ablation studies on our three designs in Tab.~\ref{tab:ablation}. We first incorporate multi-grained features for similarity computation and determine the target segments based on the similarity scores of aggregated visual features. This leads to slight improvements across various TVG subtasks~\cite{liu2024bench}, supporting our assumption that multi-grained information provides valuable cues and helps compensate for information loss. \revise{We also directly apply Savitzky-Golay smoothing~\cite{savitzky1964smoothing} to the similarity curves of the E.T.Chat baseline~\cite{liu2024bench}. However, this variant produces no changes because E.T.Chat independently determines the start and end boundaries from two peak similarity responses. Suppressing local fluctuations alone does not affect the peak as discussed in Sec.~\ref{ssec:sg_smooth}.} Next, we replace timestamp matching with event matching to better capture the semantic continuity and integrity of query events, resulting in notable performance gains, \textit{e.g.}, 6.9 on TVG and 5.3 on EVS. To further refine the final predictions, we apply the Savitzky-Golay smoothing to the similarity scores within each event. This additional step yields further performance improvements on TVG tasks, validating our hypothesis that the Savitzky-Golay smoothing contributes to more coherent and accurate event localization. \revise{However, the TAL performance remains nearly unchanged, with the full model achieving \(30.9\) compared with \(30.8\) for the baseline. TAL requires the model to recognize and enumerate all occurrences of an action in a long video, whereas E.M.Ground primarily improves the integrity and boundaries of each generated event span. It does not explicitly address missed action instances or incorrect event counts, which limits its improvement on TAL. } \revise{To further validate the effectiveness of the proposed event matching strategy, we conduct a controlled experiment between dual start-end token matching and the single \(<\!\!evt\!\!>\) token. Both variants use the same architecture, dataset, and training protocol, and differ only in the matching target. As shown in Tab.~\ref{tab:ablation}, single-token event matching achieves better performance on most tasks. This confirms that modeling the holistic event span is more effective than independently matching isolated temporal boundaries.}

\begin{figure}[t]
\centering
    \vspace{-12mm}
    \includegraphics[width=0.98\linewidth]{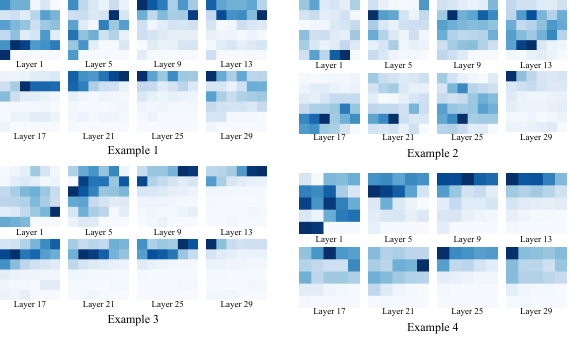}
    \vspace{-6mm}
    \caption{\revise{Visualization of attention patterns between the \(<\!\!evt\!\!>\) token and video frames across different LLM layers. Each block corresponds to the attention weight assigned to a specific frame. Darker colors indicate higher attention weights, while lighter colors indicate lower attention weights. The diverse attention patterns across layers show that different LLM layers capture complementary temporal cues, which motivates our multi-grained feature aggregation for event grounding. The quantitative effectiveness is validated by the controlled ablation in Tab.~\ref{tab:ablation}.}}
    \vspace{+2mm}
\label{fig:multi-layer}
\end{figure}

\input{tab_aggregation}

\subsection{\secondrevise{Analysis of Multi-grained Feature Aggregation}}
Beyond the quantitative results in Tab.~\ref{tab:ablation}, we visualize the attentions between $<\!\!evt\!\!>$ token and all video frames in Fig.~\ref{fig:multi-layer}. We observe that the attention patterns are very distinct across different layers, which further demonstrates our assumption that utilizing the multi-grained visual features across all layers brings richer information for temporal grounding. \secondrevise{In addition to the quantitative component ablation in Tab.~\ref{tab:ablation} and the layer-wise attention visualization in Fig.~\ref{fig:multi-layer}, we compare different strategies for aggregating frame features across LLM layers. Given the layer-wise features \(\{\mathbf{f}_{v_t}^{\,l}\}_{l=1}^{L}\), we consider element-wise maximum, element-wise median, and uniform averaging. Uniform averaging can be interpreted as the least-squares consensus of the layer-wise representations:
  \[ \overline{\mathbf{f}}_{v_t} = \arg\min_{\mathbf{z}} \sum_{l=1}^{L} \left\| \mathbf{z}-\mathbf{f}_{v_t}^{\,l} \right\|_2^2 = \frac{1}{L}\sum_{l=1}^{L}\mathbf{f}_{v_t}^{\,l}. \]
This strategy preserves contributions from all layers without introducing additional parameters or favoring a particular layer. As shown in Tab.~\ref{tab:aggregation}, uniform averaging achieves the best overall performance. Max aggregation may emphasize isolated layer-specific activations, whereas median aggregation may suppress complementary features that are strongly represented in only a subset of layers. These results support uniform averaging as a simple and effective strategy for constructing a consensus representation from multi-grained LLM features.}

\subsection{Computational Complexity and Time Cost}
E.M.Ground exhibits linear complexity with respect to the number of frame tokens $T$ when predicting the event period from the $<\!\!evt\!\!>$ token. Specifically, the token-to-frame similarity computation incurs a cost of $O(T \times D)$, and the Savitzky–Golay smoothing introduces an additional cost of $O(T \times K)$, where $D$ is the embedding dimension and $K$ is the smoothing window size. As both $D$ and $K$ are fixed constants, the overall computational complexity reduces to $O(T)$. As a result, the overall computational overhead is negligible, adding only approximately 0.3s per prediction. Moreover, the inference time of E.M.Ground (38m47s / 1{,}000 QA pairs in TVG tasks from E.T.Bench) is nearly identical to that of E.T.Chat~\cite{liu2024bench} (34m27s / 1{,}000 Question-Answer pairs). \secondrevise{To provide a more complete computational profile, we additionally measure the FLOPs and GPU memory using a representative 40-second video. Following the default 1-FPS sampling strategy, the input contains \(T=40\) frames. On a single A800 GPU with FP16 precision and a batch size of 1, E.M.Ground requires $22.8 \times 10^{12}$ FLOPs and 10.9GB of GPU memory. Its end-to-end inference time is 2.23s. These measurements demonstrate that E.M.Ground remains computationally lightweight in terms of both computation and memory usage.}

\input{tab_dvc_etbench.tex}

\input{tab_dvc_youcook2.tex}

\input{tab_vhd.tex}

\subsection{Performance on Related Tasks}~\label{ssec:related_task}
\subsubsection{DVC} In Tabs.~\ref{tab:dvc_etbench} and \ref{tab:dvc_youcook2}, we report the performance of E.M.Ground on two representative DVC benchmarks compared with recent Vid-LLMs~\cite{ren2024timechat,deng2024seq2time,guo2025vtg,maaz2023video,xu2024pllava,huang2024vtimellm,li2024llama,lin2023video,huang2024lita,liu2024bench}. The results demonstrate that E.M.Ground achieves strong performance across all metrics. Although our designs primarily aim to enhance temporal localization capabilities, E.M.Ground also shows consistent improvements in accurately describing video segments, \textit{i.e.}, SODA\_c and CIDEr on YouCook2, and F1 on E.T.Bench. We attribute this to the loss function leveraging multi-grained visual features, which provides effective supervision and benefits the model’s captioning ability as well.

\subsubsection{VHD} As shown in Tab.~\ref{tab:vhd}, E.M.Ground also demonstrates strong performance on the VHD task compared with state-of-the-art Vid-LLMs~\cite{qian2024momentor,ren2024timechat,wu2024number,guo2025vtg}. Unlike TVG and DVC, where the outputs are time segments, VHD requires predicting precise timestamps. This result supports our assumption that a unified representation can effectively benefit Vid-LLMs across diverse temporal understanding tasks.

\input{tab_mvbench.tex}

\subsection{Performance on MVBench}
Although E.M.Ground is primarily designed for TVG tasks, its well-designed modules also benefit general video understanding tasks. We report its performance on MVBench~\cite{li2024mvbench} in Tab.~\ref{tab:mvbench}, where E.M.Ground achieves competitive average results compared with other temporal-grounding Vid-LLMs with larger model sizes. Specifically, E.M.Ground surpasses the timestamp-matching Vid-LLM E.T.Chat~\cite{liu2024bench} in most cases (the $\Delta$ in the last row indicates the performance improvements), further demonstrating that our design maintains strong generalization ability without compromising the overall capability of the baseline. \secondrevise{Together with the caption generation results in Sec.~\ref{ssec:related_task}, these experiments provide task-level evidence for the compatibility between the NLL and matching objectives. Specifically, E.M.Ground improves the DVC and SLC similarity scores of E.T.Chat from 19.7 to 21.5 and from 14.6 to 16.2, respectively, while also improving the average MVBench performance from 37.1 to 42.0. These results show that the temporal supervision introduced by \(\mathcal{L}_{\mathrm{matching}}\) does not cause an observable degradation in descriptive generation or general video understanding.}

\input{tab_hpt.tex}

\subsection{Robustness against Hyperparameters}\label{ssec:hpt}

E.M.Ground incorporates several hyperparameters: the threshold $\sigma$ for selecting predicted event spans, \revise{the window size \(k\), and the polynomial order \(p\) used in the Savitzky-Golay smoothing. Although these hyperparameters need to be specified during training or inference, they do not affect the generalization ability of our method.}  \secondrevise{As shown in Tab.~\ref{tab:hpt}, the performance remains stable across a wide range of values for all these hyperparameters. We further analyze the robustness of the soft boundary labels by varying the decay factor \(\alpha\) and the boundary constraint length \(d\). The factor \(\alpha\) controls the label decay rate outside the annotated event, while \(d\) controls the temporal extent of the soft transition. A smaller \(\alpha\) or a larger \(d\) produces a slower and wider transition suitable for gradual event boundaries, whereas a larger \(\alpha\) or a smaller \(d\) produces a sharper and narrower transition suitable for abrupt boundaries. As shown in Tab.~\ref{tab:hpt}, the performance remains stable when \(\alpha\) varies from 1 to 3 and \(d\) varies from 1 to 5. Specifically, R@0.5 remains within 43.5-44.8 for different values of \(\alpha\) and within 44.1-44.8 for different values of \(d\), while mIoU varies by at most 0.3 points in both cases. These results demonstrate that E.M.Ground is robust to soft-label profiles ranging from gradual to relatively abrupt boundary transitions. We adopt \(\alpha=2\) and \(d=3\) as a balanced default setting.} Notably, we adopt the same set of hyperparameter values in all tasks and datasets. More discussions on smoothing strategy, aggregation strategy, and performance robustness are provided in the supplementary materials.

\input{tab_dynamic_threshold}

\subsection{\revise{Dynamic Thresholding Strategy}}

\revise{To further examine the effect of threshold selection, we also evaluate a dynamic thresholding strategy. Given the smoothed similarity scores \(\mathbf{s}=\{s_t\}_{t=1}^{T}\), we first normalize them within each video as \(\hat{s}_t = (s_t-\min_j s_j)/(\max_j s_j-\min_j s_j+\epsilon)\). We then define the dynamic threshold as \(\sigma_{\mathrm{dyn}}=\rho \cdot \max_t \hat{s}_t\), where \(\rho\) controls the relative threshold ratio. Since \(\max_t \hat{s}_t=1\) after normalization, \(\sigma_{\mathrm{dyn}}=\rho\). The final event span is selected as the contiguous region around the peak frame whose normalized similarity satisfies \(\hat{s}_t \geq \sigma_{\mathrm{dyn}}\). As shown in Tab.~\ref{tab:threshold_dynamic}, the dynamic strategy achieves performance comparable to the fixed threshold, suggesting that E.M.Ground is not sensitive to the exact threshold form. This is because the learned event-response curve already separates event frames from non-event frames, making the threshold mainly serve as a simple foreground selection criterion.}

\input{tab_smoothing}

\subsection{\revise{Discussion on Different Smoothing Strategies}}

\revise{We compare Savitzky-Golay smoothing with moving average, exponential moving average, and Gaussian filtering. For a fair comparison, the moving-average, Gaussian, and Savitzky-Golay filters use a window size of \(k=5\). The Gaussian standard deviation is set to \(\sigma_g=1.0\), and the exponential moving average uses a span of \(5\). As shown in Tab.~\ref{tab:smooth_main}, Savitzky-Golay smoothing achieves the best overall performance. Moving average and Gaussian filtering may attenuate short responses or merge adjacent events, while exponential moving average may introduce temporal lag. In contrast, Savitzky-Golay smoothing better preserves peak positions and event boundaries while suppressing local noise.}

\input{tab_pooling_evt}

\subsection{\secondrevise{Comparison with Non-parametric Pooling Baselines}}

\secondrevise{To determine whether a dedicated learnable event token is necessary, we compare the $<\!\!evt\!\!>$ representation with two non-parametric alternatives based on query mean pooling and query max pooling. Specifically, we apply the pooling operations to the hidden states corresponding only to the natural-language event-description tokens. Video tokens, generic instruction tokens, answer-prefix tokens, and the $<\!\!evt\!\!>$ token are excluded from pooling. The pooled feature replaces \(f_{\mathrm{evt}}^{L-1}\) in Eq.~3 and is passed through the same event projector, while all other training and inference settings remain unchanged. As shown in Tab.~\ref{tab:pooling_evt}, both pooling baselines perform substantially worse than the learnable $<\!\!evt\!\!>$ representation. The pooling baselines construct an event representation by applying a fixed aggregation operation to token-level query features and do not provide a dedicated representation position for the complete event. In contrast, $<\!\!evt\!\!>$ serves explicitly as an event-level representation and is optimized as a whole to match all frames within the ground-truth event span. These results demonstrate that the dedicated $<\!\!evt\!\!>$ representation is more effective than fixed query-token pooling for event-to-frame matching.}

\section{Conclusion}
In this work, we propose E.M.Ground, a novel method for Temporal Video Grounding (TVG) based on Video Large Language Models (Vid-LLMs). By reformulating timestamp prediction as an event matching task with a unified token representation, E.M.Ground overcomes the limitations of conventional timestamp matching methods in capturing the semantic continuity and integrity of query events. To further enhance temporal grounding, we introduce multi-grained visual feature aggregation and apply Savitzky-Golay smoothing for prediction refinement.
\secondrevise{Extensive experiments on multiple TVG benchmarks demonstrate that our approach achieves consistent improvements over previous methods, with the most pronounced gains on single-event grounding and boundary-integrity tasks.} We hope it could provide new perspectives on precise temporal localization with Vid-LLMs.

\section*{Declaration of Competing Interests}
The authors declare that they have no known competing financial interests or personal relationships that could have appeared to influence the work in this paper.

\section*{CRediT Authorship Contribution Statement}
Jiahao Nie: Writing - Original draft, Software, Methodology. Wenbin An: Writing - Polished version, Software, Resources. Gongjie Zhang: Conceptualization, Formal analysis. Yicheng Xu: Resources, Conceptualization, Formal analysis. Yap-Peng Tan: Supervision. Alex C. Kot: Supervision. Shijian
Lu: Supervision, Project administration, Funding acquisition.

\section*{Acknowledgement}
This study is supported by the Interdisciplinary Graduate Programme, Nanyang Technological University, and the Ministry of Education Singapore, under the Tier-2 scheme with project number MOE-T2EP20123-0003.

%% file: tab_intermediate_distractor.tex
\begin{table}[t]
    \centering
    \small
    \renewcommand\arraystretch{0.9}
    \setlength{\tabcolsep}{5pt}
    
    \vspace{-6mm}
    \caption{\secondrevise{Performance after relocating the original event-ending frame on the E.T.Bench TVG subtask. ``New-end selection'' denotes selecting \(v_{e-1}\), while ``relocated-end selection'' denotes incorrectly following the relocated \(v_e\).}}
    \vspace{2mm}

    \scalebox{0.86}{
    \begin{tabular}{l||cc|cc}
        \toprule[1pt]
        \secondrevise{Model}
        & \secondrevise{Original F1 (\(\uparrow\))} & \secondrevise{F1 after relocation (\(\uparrow\))} & \secondrevise{New-end selection (\(\uparrow\))} & \secondrevise{Relocated-end selection (\(\downarrow\))} \\ \hline\hline
        \secondrevise{E.T.Chat} & \secondrevise{38.6} & \secondrevise{30.7} & \secondrevise{22.6\%} & \secondrevise{62.4\%} \\
        \secondrevise{\textbf{Ours}} & \secondrevise{\textbf{52.0}} & \secondrevise{\textbf{49.9}} & \secondrevise{\textbf{68.7\%}} & \secondrevise{\textbf{11.8\%}} \\
        \bottomrule[1pt]
      \end{tabular}
    }
    \label{tab:intermediate_distractor}
  \end{table}

%% file: tab_ablation.tex

\begin{table}[t]
    \renewcommand\arraystretch{0.9}
    \centering
    \vspace{-12mm}
    \caption{Ablation studies for technical designs.}
    \vspace{+2mm}
    
    \small
    \setlength{\tabcolsep}{12pt}
    \scalebox{0.88}{
    \begin{tabular}{l|c|c|c|c}
    \toprule[1pt]
    \multicolumn{5}{c}{\textit{\textbf{E.T.Bench-Grounding}}}\\\hline
    Setup& TVG& EPM& TAL& EVS\\\hline\hline
    \revise{Baseline (E.T.Chat, dual token matching)}  & 38.6& 10.2& 30.8& 25.4\\
    \revise{Baseline + Multi-grained feature}  & \revise{41.3}& \revise{10.8}& \revise{30.5}& \revise{26.5}\\
    \revise{Single \(<\!\!evt\!\!>\) token matching} & \revise{45.1} & \revise{11.8} & \revise{30.8} & \revise{30.6} \\
    \revise{Baseline + Savitzky-Golay Smoothing}  & \revise{38.6}& \revise{10.2}& \revise{30.8}& \revise{25.4}\\ \hline\hline
    Baseline (E.T.Chat, dual token matching)  & 38.6& 10.2& 30.8& 25.4\\
    + Multi-grained feature  & 41.3& 10.8& 30.5& 26.5\\
    + Event matching  & 48.2& 12.8& 30.8& 31.8\\
    \cellcolor{mycolor}\textbf{+ Savitzky-Golay Smoothing}      & \cellcolor{mycolor}\textbf{52.0}& \cellcolor{mycolor}\textbf{14.0}& \cellcolor{mycolor}\textbf{30.9}& \cellcolor{mycolor}\textbf{32.8}\\
    \bottomrule[1pt]
    \end{tabular}
    }

    \label{tab:ablation}
\end{table}

%% file: tab_aggregation.tex
\begin{table}[t]
\centering
\small
\renewcommand\arraystretch{0.9}
\setlength{\tabcolsep}{8pt}
\vspace{-4mm}
\caption{\secondrevise{Performance comparison  across different LLM-layer aggregation strategies.}}
\vspace{+2mm}
\label{tab:aggregation}
\setlength{\tabcolsep}{15pt}
\scalebox{0.86}{
    \begin{tabular}{l||cc}
    \toprule[1pt]
    \multicolumn{3}{c}{\textit{\textbf{\secondrevise{Charades-STA}}}}\\\hline
    \secondrevise{Aggregation strategy} & \secondrevise{R@0.5} & \secondrevise{mIoU} \\\hline\hline
    \secondrevise{max} & \secondrevise{41.6} & \secondrevise{40.7}\\
    \secondrevise{median} & \secondrevise{42.5} & \secondrevise{42.1} \\
    \cellcolor{mycolor}\secondrevise{\textbf{average}} & \cellcolor{mycolor}\secondrevise{\textbf{44.8}} & \cellcolor{mycolor}\secondrevise{\textbf{43.3}}\\ 
    \bottomrule[1pt]
\end{tabular}
}
\end{table}

%% file: tab_dvc_etbench.tex
\begin{table}[t]
    \renewcommand\arraystretch{0.9}
    \centering
    \vspace{-12mm}
    
    \caption{Zero-shot performance on the Dense Video Captioning subtasks of E.T.Bench~\cite{liu2024bench}. The best results for each metric are highlighted in \textbf{bold} font.}
    \vspace{+2mm}
    \small
    \setlength{\tabcolsep}{18pt}
    \scalebox{0.8}{
    \begin{tabular}{l|c|c|c|c}
    \toprule[1pt]
    \multicolumn{5}{c}{\textit{\textbf{E.T.Bench-Captioning}}}\\\hline
    \multirow{2}*{Methods}& \multicolumn{2}{c|}{DVC}& \multicolumn{2}{c}{SLC}\\\cline{2-5}
    ~& F1& Sim& F1& Sim\\\hline\hline
    Video-ChatGPT(7B)~\cite{maaz2023video} & 8.8& 11.3& 5.7& 10.2\\
    PLLaVA (7B)~\cite{xu2024pllava}      & 13.3& 10.6& 9.7& 11.8\\
    VTimeLLM (7B)~\cite{huang2024vtimellm}    & 12.4& 13.1& 8.7& 6.4\\
    TimeChat (7B)~\cite{ren2024timechat}    & 16.6& 12.5& 5.6& 9.2\\
    LLaMA-VID (7B)~\cite{li2024llama}   & 27.1& 12.6& 5.2& 11.1\\  
    Video-LLaVA (7B)~\cite{lin2023video} & 28.0& 15.0& 0.9& 8.3\\
    E.T.Chat (3.8B)~\cite{liu2024bench}  & 38.4& 19.7& 24.4& 14.6\\
    LITA (7B)~\cite{huang2024lita}        & 39.7& 17.2& 21.0& 12.2\\
    VTG-LLM (7B)~\cite{guo2025vtg}     & 40.2& 18.6& 20.8& 14.4\\
    \cellcolor{mycolor}\textbf{Ours (3.8B)}      & \cellcolor{mycolor}\textbf{41.6}& \cellcolor{mycolor}\textbf{21.5}& \cellcolor{mycolor}\textbf{25.6}& \cellcolor{mycolor}\textbf{16.2}\\
    \bottomrule[1pt]
    \end{tabular}
    }
    
    \label{tab:dvc_etbench}
\end{table}

%% file: tab_dvc_youcook2.tex
\begin{table}[ht]
    \renewcommand\arraystretch{0.9}
    \centering
    \vspace{-2mm}
    
    \caption{Zero-shot performance on the YouCook2 dataset~\cite{zhou2018towards}, designed for the Dense Video Captioning task. The best results for each metric are highlighted in \textbf{bold} font.}
    \vspace{+2mm}
    \small
    \setlength{\tabcolsep}{12pt}
    \scalebox{0.82}{
    \begin{tabular}{l|cc|c}
    \toprule[1pt]
    \multicolumn{4}{c}{\textit{\textbf{YouCook2}}}\\\hline
    \multirow{2}*{Methods}& \multicolumn{2}{c|}{Caption}& Time\\\cline{2-4}
    ~& SODA\_c& CIDEr& F1\\\hline\hline
    TimeChat (7B)~\cite{ren2024timechat} & 1.2& 3.4& 12.6\\
    Seq2Time (7B)~\cite{deng2024seq2time} & 1.3& 4.2& 16.2\\
    E.T.Chat(3.8B)~\cite{liu2024bench} & 1.5& 4.7& 15.1\\
    VTG-LLM (7B)~\cite{guo2025vtg}  & 1.5& 5.0& 17.5\\
    \cellcolor{mycolor}\textbf{Ours (3.8B)}   & \cellcolor{mycolor}\textbf{1.6}& \cellcolor{mycolor}\textbf{5.1}& \cellcolor{mycolor}\textbf{20.1}\\
    \bottomrule[1pt]
    \end{tabular}
    }

    \vspace{+2mm}
    \label{tab:dvc_youcook2}
\end{table}

%% file: tab_vhd.tex
\begin{table}[h]
    \renewcommand\arraystretch{0.9}
    \centering
    \vspace{-12mm}
    \caption{Zero-shot performance on the QVHighlights dataset~\cite{lei2021detecting}, designed for the Video Highlight Detection task. The best results for each metric are highlighted by \textbf{bold} font.}
    \vspace{+2mm}
    \small
    \setlength{\tabcolsep}{22pt}
    \scalebox{0.82}{
    \begin{tabular}{l|cc}
    \toprule[1pt]
    \multicolumn{3}{c}{\textit{\textbf{QVHighlights}}}\\\hline
    Methods& mAP& HIT@1\\\hline\hline
    Momentor (7B)~\cite{qian2024momentor} & 7.6& -\\
    TimeChat (7B)~\cite{ren2024timechat} & 14.5& 23.9\\
    NumPro (7B)~\cite{wu2024number}   & 15.3& 24.3\\
    VTG-LLM (7B)~\cite{guo2025vtg}  & 16.3& 33.5\\
    E.T.Chat (3.8B)~\cite{liu2024bench} & 20.0& 44.8\\
    \cellcolor{mycolor}\textbf{Ours (3.8B)}   & \cellcolor{mycolor}\textbf{21.5}& \cellcolor{mycolor}\textbf{52.5}\\
    \bottomrule[1pt]
    \end{tabular}
    }

    \label{tab:vhd}
\end{table}

%% file: tab_mvbench.tex
\begin{table*}[t]
\renewcommand\arraystretch{0.9}
\centering
\vspace{-2mm}

\caption{Performance on the MVBench~\cite{li2024mvbench}. The best results are in \textbf{bold}, and $\Delta$ indicates improvement over E.T.Chat~\cite{liu2024bench}.}
\vspace{+2mm}

\setlength{\tabcolsep}{6pt}
\small

\scalebox{0.8}{
\begin{tabular}{l|cccccccccc}
\toprule[1pt]
\multicolumn{11}{c}{\textbf{MVBench (Part 1)}}\\
\hline
Methods & Avg. & AA & AC & AL & AP & AS & CO & CI & EN & ER \\
\hline\hline
VideoLLaMA & 34.1 & 51.0 & 34.0 & 22.5 & 25.5 & 27.5 & 40.0 & \textbf{37.0} & 30.0 & 21.0 \\
VideoChat & 35.5 & 56.0 & 35.0 & 27.0 & 26.5 & 33.5 & \textbf{41.0} & 36.0 & 23.5 & 23.5 \\
TimeChat & 38.5 & 61.0 & 34.0 & 26.5 & \textbf{36.0} & \textbf{40.5} & 36.0 & 35.0 & 29.0 & 35.0 \\
E.T.Chat & 37.1 & 59.5 & 28.0 & 38.5 & 33.5 & 34.5 & 31.0 & 22.5 & 27.0 & 49.0 \\
\rowcolor{mycolor}\textbf{Ours} & \textbf{42.0} & \textbf{73.0} & \textbf{50.0} & \textbf{39.0} & 33.5 & 35.5 & 39.0 & 32.0 & \textbf{32.5} & \textbf{51.0} \\
\rowcolor{mycolor}\textbf{$\Delta$} & \textbf{+4.9} & \textbf{+13.5} & \textbf{+22.0} & \textbf{+0.5} & 0.0 & \textbf{+1.0} & \textbf{+8.0} & \textbf{+9.5} & \textbf{+5.5} & \textbf{+2.0} \\
\bottomrule[1pt]
\end{tabular}
}

\vspace{1mm}

\setlength{\tabcolsep}{5pt}
\scalebox{0.8}{
\begin{tabular}{l|ccccccccccc}
\toprule[1pt]
\multicolumn{12}{c}{\textbf{MVBench (Part 2)}}\\
\hline
Methods & FA & FP & MA & MC & MD & OE & OI & OS & ST & SC & UA \\
\hline\hline
VideoLLaMA & 29.0 & 32.5 & 32.5 & 22.5 & 22.5 & 48.0 & 40.5 & \textbf{38.0} & 43.0 & 45.5 & 39.0 \\
VideoChat & \textbf{33.5} & 26.5 & 42.5 & 20.5 & 25.5 & 53.0 & 40.5 & 30.0 & 48.5 & \textbf{46.0} & 40.5 \\
TimeChat & 32.5 & 36.5 & \textbf{43.5} & 20.0 & 19.5 & \textbf{53.5} & \textbf{41.5} & 29.0 & 66.5 & 42.0 & 53.0 \\
E.T.Chat & 29.5 & 38.0 & 36.5 & 26.0 & 24.5 & 51.0 & 33.0 & 32.5 & 52.5 & 37.5 & \textbf{60.0} \\
\rowcolor{mycolor}\textbf{Ours} & 30.5 & \textbf{39.5} & 37.0 & \textbf{33.5} & \textbf{26.0} & 48.0 & 40.0 & 34.5 & \textbf{67.5} & 40.5 & \textbf{60.0} \\
\rowcolor{mycolor}\textbf{$\Delta$} & \textbf{+1.0} & \textbf{+1.5} & \textbf{+0.5} & \textbf{+7.5} & \textbf{+1.5} & -3.0 & \textbf{+7.0} & \textbf{+2.0} & \textbf{+15.0} & \textbf{+3.0} & 0.0 \\
\bottomrule[1pt]
\end{tabular}
}

\label{tab:mvbench}
\end{table*}

%% file: tab_hpt.tex
\begin{table}[t]      
  \centering            
  \small            
  \renewcommand\arraystretch{0.9}             
  \setlength{\tabcolsep}{4pt}        

  \vspace{-12mm}
  
  \caption{\revise{E.M.Ground exhibits stable performance across a wide range of hyperparameters.}}    \vspace{+2mm}                                           \scalebox{0.86}{        
  \begin{tabular}{c|cc||c|cc||c|cc}
  \toprule[1pt]
  \multicolumn{3}{c||}{$\sigma$} & \multicolumn{3}{c||}{$k$} &  \multicolumn{3}{c}{\revise{$p$}} \\\cline{1-9}
  Value & R@0.5 & mIoU & Value & R@0.5 & mIoU & \revise{Value} & \revise{R@0.5} & \revise{mIoU} \\\hline\hline
  1e-4 & 43.9 & 43.3 & 3 & 43.8 & 43.3 & \revise{1} & \revise{44.3} & \revise{43.2} \\
  8e-5 & 44.0 & 43.4 & & & & & & \\
  5e-5 & 44.0 & 43.6 & \cellcolor{mycolor}\textbf{5} & \cellcolor{mycolor}\textbf{44.8} & \cellcolor{mycolor}43.3 & \cellcolor{mycolor}\textbf{\revise{2}} & \cellcolor{mycolor}\textbf{\revise{44.8}} & \cellcolor{mycolor}\revise{43.3} \\
  3e-5 & 43.8 & \textbf{43.9} & & & & & & \\
  \cellcolor{mycolor}\textbf{1e-5} & \cellcolor{mycolor}\textbf{44.8} & \cellcolor{mycolor}43.3 & 7 & 44.0 & \textbf{43.6} &
  \revise{3} & \revise{44.5} & \revise{43.4} \\
  8e-6 & 44.2 & 43.4 & & & & & & \\
  5e-6 & 42.2 & 42.8 & 9 & 42.8 & 43.1 & \revise{4} & \revise{44.0} & \revise{43.0} \\
  \bottomrule[1pt]
  \end{tabular}
  }               

  \setlength{\tabcolsep}{12pt} 
  
  \scalebox{0.86}{
  \begin{tabular}{c|cc||c|cc}
  \toprule[1pt]
  \multicolumn{3}{c||}{\(\alpha\)} & \multicolumn{3}{c}{\revise{\(d\)}} \\\cline{1-6}
  Value & R@0.5 & mIoU & \revise{Value} & \revise{R@0.5} & \revise{mIoU} \\\hline\hline
  1 & 43.5 & 43.6 & \revise{1} & \revise{44.1} & \revise{43.1} \\
  & & & \revise{2} & \revise{44.5} & \revise{43.2} \\
  \cellcolor{mycolor}\textbf{2} & \cellcolor{mycolor}\textbf{44.8} & \cellcolor{mycolor}43.3 & \cellcolor{mycolor}\textbf{\revise{3}} & \cellcolor{mycolor}\textbf{\revise{44.8}} & \cellcolor{mycolor}\revise{43.3} \\
  & & & \revise{4} & \revise{44.6} & \textbf{\revise{43.4}} \\
  3 & 44.2 & 43.4 & \revise{5} & \revise{44.3} & \revise{43.2} \\
  \bottomrule[1pt]
  \end{tabular}
  }

\label{tab:hpt}
\end{table}

%% file: tab_dynamic_threshold.tex
\begin{table}[t]                                              
  \centering  
  \vspace{-4mm}
  
  \caption{\revise{Comparison between fixed and adaptive thresholding strategies. The adaptive results are close to the fixed-threshold results, suggesting that E.M.Ground is not sensitive to the threshold value.}}
  \vspace{+4mm}
  \label{tab:threshold_dynamic}
  \small
  \renewcommand\arraystretch{0.9}
  \setlength{\tabcolsep}{9pt}

  \scalebox{0.92}{
  \begin{tabular}{ccc||ccc}
    \toprule
    \multicolumn{3}{c||}{Fixed threshold} &
    \multicolumn{3}{c}{\revise{Adaptive threshold}} \\
    \hline
    Value & R@0.5 & mIoU & \revise{Value} & \revise{R@0.5} & \revise{mIoU} \\
    \hline\hline
    \(\sigma = 1e-4\) & 43.9 & 43.3 &
      \revise{\(\rho = 0.1\)} & \revise{44.6} & \revise{43.5} \\
      \(\sigma = 8e-5\) & 44.0 & 43.4 &
       &  &  \\
      \(\sigma = 5e-5\) & 44.0 & 43.6 &
      \revise{\(\rho = 0.2\)} & \revise{44.4} & \revise{43.4} \\
      \(\sigma = 3e-5\) & 43.8 & 43.9 &
       &  &  \\
      \(\sigma = 1e-5\) & 44.8 & 43.3 &
      \revise{\(\rho = 0.3\)} & \revise{43.9} & \revise{43.1} \\
      \(\sigma = 8e-6\) & 44.2 & 43.4 &
       &  &  \\
    \bottomrule
    \end{tabular}
    }
  \end{table}

%% file: tab_smoothing.tex
\begin{table}[t]
\renewcommand{\arraystretch}{0.9}
\centering

\small
\vspace{-12mm}

\caption{\revise{Performance comparison of different smoothing strategies.}} \label{tab:smooth_main}
\vspace{+2mm}

\setlength{\tabcolsep}{16pt}

\scalebox{0.9}{
\begin{tabular}{l|cc}
\toprule[1pt]
Smoothing strategy & R@0.5 & mIoU \\\hline\hline
Moving average (\(k=5\)) & 40.6 & 40.8 \\
Exponential moving average (span \(=5\)) & 41.0 & 41.5 \\
\revise{Gaussian filter (\(k=5,\sigma_g=1.0\))} & \revise{42.3} & \revise{42.2} \\
\cellcolor{mycolor}\textbf{Savitzky-Golay (\(k=5\))} & \cellcolor{mycolor}\textbf{44.8} & \cellcolor{mycolor}\textbf{43.3} \\
\bottomrule[1pt]

\end{tabular}
}
\end{table}

%% file: tab_pooling_evt.tex
\begin{table}[t]
\centering
\small
\renewcommand\arraystretch{0.9}
\setlength{\tabcolsep}{14pt}
\vspace{-2mm}
\caption{\secondrevise{Comparison between non-parametric query-token pooling and the $<\!\!evt\!\!>$
representation.}}
\vspace{+2mm}

\scalebox{0.88}{
\begin{tabular}{l||cc|cc}
  \toprule[1pt]
  \multirow{2}*{Event representation} & \multicolumn{2}{c|}{Charades-STA} & \multicolumn{2}{c}{E.T.Bench} \\ \cline{2-5}
  & R@0.5 & mIoU & TVG & EPM \\ \hline\hline
  \secondrevise{Query mean pooling} & \secondrevise{26.7} & \secondrevise{29.5} & \secondrevise{31.8} & \secondrevise{8.3} \\
  \secondrevise{Query max pooling} & \secondrevise{22.4} & \secondrevise{26.7} & \secondrevise{27.5} & \secondrevise{7.0} \\
  \cellcolor{mycolor}\textbf{Learnable $<\!\!evt\!\!>$ token} & \cellcolor{mycolor}\textbf{44.8} & \cellcolor{mycolor}\textbf{43.3} & \cellcolor{mycolor}\textbf{52.0} & \cellcolor{mycolor}\textbf{14.0} \\
  \bottomrule[1pt]
\end{tabular}
}

\label{tab:pooling_evt}
\end{table}

%% file: main.bib
@String(CVPR= {IEEE Conf. Comput. Vis. Pattern Recog.})

@String(ICCV= {Int. Conf. Comput. Vis.})

@String(ECCV= {Eur. Conf. Comput. Vis.})

@String(ACMMM= {ACM Int. Conf. Multimedia})

@String(ICASSP=	{ICASSP})

@String(ICLR = {Int. Conf. Learn. Represent.})

@String(AAAI = {AAAI})

@String(CVPRW= {IEEE Conf. Comput. Vis. Pattern Recog. Worksh.})

@String(CVPR  = {CVPR})

@String(ICCV  = {ICCV})

@String(ECCV  = {ECCV})

@String(ACMMM = {ACM MM})

@String(ICLR  = {ICLR})

@String(CVPRW= {CVPRW})

@inproceedings{guo2025vtg,
  title={Vtg-llm: Integrating timestamp knowledge into video llms for enhanced video temporal grounding},
  author={Guo, Yongxin and Liu, Jingyu and Li, Mingda and Cheng, Dingxin and Tang, Xiaoying and Sui, Dianbo and Liu, Qingbin and Chen, Xi and Zhao, Kevin},
  booktitle={AAAI},
  year={2025}
}

@inproceedings{guo2024trace,
  title={Trace: Temporal grounding video llm via causal event modeling},
  author={Guo, Yongxin and Liu, Jingyu and Li, Mingda and Liu, Qingbin and Chen, Xi and Tang, Xiaoying},
  booktitle={ICLR},
  year={2025}
}

@article{wang2024hawkeye,
  title={Hawkeye: Training video-text llms for grounding text in videos},
  author={Wang, Yueqian and Meng, Xiaojun and Liang, Jianxin and Wang, Yuxuan and Liu, Qun and Zhao, Dongyan},
  journal={arXiv preprint arXiv:2403.10228},
  year={2024}
}

@inproceedings{ren2024timechat,
  title={Timechat: A time-sensitive multimodal large language model for long video understanding},
  author={Ren, Shuhuai and Yao, Linli and Li, Shicheng and Sun, Xu and Hou, Lu},
  booktitle={CVPR},
  year={2024}
}

@inproceedings{huang2024vtimellm,
  title={Vtimellm: Empower llm to grasp video moments},
  author={Huang, Bin and Wang, Xin and Chen, Hong and Song, Zihan and Zhu, Wenwu},
  booktitle={CVPR},
  year={2024}
}

@inproceedings{huang2024lita,
  title={Lita: Language instructed temporal-localization assistant},
  author={Huang, De-An and Liao, Shijia and Radhakrishnan, Subhashree and Yin, Hongxu and Molchanov, Pavlo and Yu, Zhiding and Kautz, Jan},
  booktitle={ECCV},
  year={2024}
}

@inproceedings{qian2024momentor,
  title={Momentor: Advancing video large language model with fine-grained temporal reasoning},
  author={Qian, Long and Li, Juncheng and Wu, Yu and Ye, Yaobo and Fei, Hao and Chua, Tat-Seng and Zhuang, Yueting and Tang, Siliang},
  booktitle={ICML},
  year={2024}
}

@inproceedings{qu2024chatvtg,
  title={Chatvtg: Video temporal grounding via chat with video dialogue large language models},
  author={Qu, Mengxue and Chen, Xiaodong and Liu, Wu and Li, Alicia and Zhao, Yao},
  booktitle={CVPRW},
  year={2024}
}

@inproceedings{wu2024number,
  title={Number it: Temporal Grounding Videos like Flipping Manga},
  author={Wu, Yongliang and Hu, Xinting and Sun, Yuyang and Zhou, Yizhou and Zhu, Wenbo and Rao, Fengyun and Schiele, Bernt and Yang, Xu},
  booktitle={CVPR},
  year={2025}
}

@inproceedings{deng2024seq2time,
  title={Seq2Time: Sequential Knowledge Transfer for Video LLM Temporal Grounding},
  author={Deng, Andong and Gao, Zhongpai and Choudhuri, Anwesa and Planche, Benjamin and Zheng, Meng and Wang, Bin and Chen, Terrence and Chen, Chen and Wu, Ziyan},
  booktitle={CVPR},
  year={2025}
}

@inproceedings{liu2024bench,
  title={Et bench: Towards open-ended event-level video-language understanding},
  author={Liu, Ye and Ma, Zongyang and Qi, Zhongang and Wu, Yang and Shan, Ying and Chen, Chang W},
  booktitle={NeurIPS},
  year={2024}
}

@article{xu2024pllava,
  title={Pllava: Parameter-free llava extension from images to videos for video dense captioning},
  author={Xu, Lin and Zhao, Yilin and Zhou, Daquan and Lin, Zhijie and Ng, See Kiong and Feng, Jiashi},
  journal={arXiv preprint arXiv:2404.16994},
  year={2024}
}

@inproceedings{lin2023video,
  title={Video-llava: Learning united visual representation by alignment before projection},
  author={Lin, Bin and Ye, Yang and Zhu, Bin and Cui, Jiaxi and Ning, Munan and Jin, Peng and Yuan, Li},
  booktitle={EMNLP},
  year={2024}
}

@inproceedings{wang2023internvid,
  title={Internvid: A large-scale video-text dataset for multimodal understanding and generation},
  author={Wang, Yi and He, Yinan and Li, Yizhuo and Li, Kunchang and Yu, Jiashuo and Ma, Xin and Li, Xinhao and Chen, Guo and Chen, Xinyuan and Wang, Yaohui and others},
  booktitle={ICLR},
  year={2023}
}

@article{zhang2021multi,
  title={Multi-scale 2d temporal adjacency networks for moment localization with natural language},
  author={Zhang, Songyang and Peng, Houwen and Fu, Jianlong and Lu, Yijuan and Luo, Jiebo},
  journal={TPAMI},
  year={2021}
}

@inproceedings{chen2025longvila,
  title={Longvila: Scaling long-context visual language models for long videos},
  author={Chen, Yukang and Xue, Fuzhao and Li, Dacheng and Hu, Qinghao and Zhu, Ligeng and Li, Xiuyu and Fang, Yunhao and Tang, Haotian and Yang, Shang and Liu, Zhijian and others},
  booktitle={ICLR},
  year={2025}
}

@inproceedings{di2024grounded,
  title={Grounded question-answering in long egocentric videos},
  author={Di, Shangzhe and Xie, Weidi},
  booktitle={CVPR},
  year={2024}
}

@inproceedings{wang2025time,
  title={Time-R1: Post-Training Large Vision Language Model for Temporal Video Grounding},
  author={Wang, Ye and Wang, Ziheng and Xu, Boshen and Du, Yang and Lin, Kejun and Xiao, Zihan and Yue, Zihao and Ju, Jianzhong and Zhang, Liang and Yang, Dingyi and others},
  booktitle={NeurIPS},
  year={2025}
}

@inproceedings{li2024mvbench,
  title={Mvbench: A comprehensive multi-modal video understanding benchmark},
  author={Li, Kunchang and Wang, Yali and He, Yinan and Li, Yizhuo and Wang, Yi and Liu, Yi and Wang, Zun and Xu, Jilan and Chen, Guo and Luo, Ping and others},
  booktitle={CVPR},
  year={2024}
}

@inproceedings{fu2024video,
  title={Video-mme: The first-ever comprehensive evaluation benchmark of multi-modal llms in video analysis},
  author={Fu, Chaoyou and Dai, Yuhan and Luo, Yongdong and Li, Lei and Ren, Shuhuai and Zhang, Renrui and Wang, Zihan and Zhou, Chenyu and Shen, Yunhang and Zhang, Mengdan and others},
  booktitle={CVPR},
  year={2025}
}

@inproceedings{wang2022negative,
  title={Negative sample matters: A renaissance of metric learning for temporal grounding},
  author={Wang, Zhenzhi and Wang, Limin and Wu, Tao and Li, Tianhao and Wu, Gangshan},
  booktitle={AAAI},
  year={2022}
}

@inproceedings{zhang2020span,
  title={Span-based localizing network for natural language video localization},
  author={Zhang, Hao and Sun, Aixin and Jing, Wei and Zhou, Joey Tianyi},
  booktitle={ACL},
  year={2020}
}

@inproceedings{dziri2023faith,
  title={Faith and fate: Limits of transformers on compositionality},
  author={Dziri, Nouha and Lu, Ximing and Sclar, Melanie and Li, Xiang Lorraine and Jiang, Liwei and Lin, Bill Yuchen and Welleck, Sean and West, Peter and Bhagavatula, Chandra and Le Bras, Ronan and others},
  booktitle={NeurIPS},
  year={2023}
}

@inproceedings{song2025symmetrical,
  title={Symmetrical Awareness Generation for Pelvic Image Segmentation},
  author={Song, Yize and Chen, Yunqing and Wang, Zhou and Chen, Cheng and Xiao, Ruoxiu},
  booktitle={ACMMM},
  year={2025}
}

@article{chen2022generative,
  title={Generative consistency for semi-supervised cerebrovascular segmentation from TOF-MRA},
  author={Chen, Cheng and Zhou, Kangneng and Wang, Zhiliang and Xiao, Ruoxiu},
  journal={TMI},
  year={2022}
}

@inproceedings{carion2025sam3,
  title={Sam 3: Segment anything with concepts},
  author={Carion, Nicolas and Gustafson, Laura and Hu, Yuan-Ting and Debnath, Shoubhik and Hu, Ronghang and Suris, Didac and Ryali, Chaitanya and Alwala, Kalyan Vasudev and Khedr, Haitham and Huang, Andrew and others},
  booktitle={ICLR},
  year={2026}
}

@inproceedings{lin2025multi,
  title={Multi-Layer Visual Feature Fusion in Multimodal LLMs: Methods, Analysis, and Best Practices},
  author={Lin, Junyan and Chen, Haoran and Fan, Yue and Fan, Yingqi and Jin, Xin and Su, Hui and Fu, Jinlan and Shen, Xiaoyu},
  booktitle={CVPR},
  year={2025}
}

@article{yao2024dense,
  title={Dense connector for mllms},
  author={Yao, Huanjin and Wu, Wenhao and Yang, Taojiannan and Song, YuXin and Zhang, Mengxi and Feng, Haocheng and Sun, Yifan and Li, Zhiheng and Ouyang, Wanli and Wang, Jingdong},
  journal={NeurIPS},
  year={2024}
}

@inproceedings{azad2025hierarq,
  title={HierarQ: Task-Aware Hierarchical Q-Former for Enhanced Video Understanding},
  author={Azad, Shehreen and Vineet, Vibhav and Rawat, Yogesh Singh},
  booktitle={CVPR},
  year={2025}
}

@inproceedings{he2024multi,
  title={Multi-modal instruction tuned llms with fine-grained visual perception},
  author={He, Junwen and Wang, Yifan and Wang, Lijun and Lu, Huchuan and He, Jun-Yan and Lan, Jin-Peng and Luo, Bin and Xie, Xuansong},
  booktitle={CVPR},
  year={2024}
}

@article{savitzky1964smoothing,
  title={Smoothing and differentiation of data by simplified least squares procedures.},
  author={Savitzky, Abraham and Golay, Marcel JE},
  journal={Analytical chemistry},
  year={1964}
}

@book{hamming1998digital,
  title={Digital filters},
  author={Hamming, Richard Wesley},
  year={1998},
  publisher={Courier Corporation}
}

@inproceedings{pandia2010motion,
  title={Motion artifact cancellation to obtain heart sounds from a single chest-worn accelerometer},
  author={Pandia, Keya and Ravindran, Sourabh and Cole, Randy and Kovacs, Gregory and Giovangrandi, Laurent},
  booktitle={ICASSP},
  year={2010}
}

@incollection{persson2003smoothing,
  title={Smoothing by savitzky-golay and legendre filters},
  author={Persson, Per-Olof and Strang, Gilbert},
  booktitle={Mathematical systems theory in biology, communications, computation, and finance},
  year={2003},
  publisher={Springer}
}

@inproceedings{dosovitskiy2020image,
  title={An image is worth 16x16 words: Transformers for image recognition at scale},
  author={Dosovitskiy, Alexey and Beyer, Lucas and Kolesnikov, Alexander and Weissenborn, Dirk and Zhai, Xiaohua and Unterthiner, Thomas and Dehghani, Mostafa and Minderer, Matthias and Heigold, Georg and Gelly, Sylvain and others},
  booktitle={ICLR},
  year={2021}
}

@inproceedings{fang2023eva,
  title={Eva: Exploring the limits of masked visual representation learning at scale},
  author={Fang, Yuxin and Wang, Wen and Xie, Binhui and Sun, Quan and Wu, Ledell and Wang, Xinggang and Huang, Tiejun and Wang, Xinlong and Cao, Yue},
  booktitle={CVPR},
  year={2023}
}

@inproceedings{dai2023instructblip,
  title={InstructBLIP: Towards General-purpose Vision-Language Models with Instruction Tuning}, 
  author={Wenliang Dai and Junnan Li and Dongxu Li and Anthony Meng Huat Tiong and Junqi Zhao and Weisheng Wang and Boyang Li and Pascale Fung and Steven Hoi},
  booktitle={NeurIPS},
  year={2023}
}

@inproceedings{li2023blip,
  title={Blip-2: Bootstrapping language-image pre-training with frozen image encoders and large language models},
  author={Li, Junnan and Li, Dongxu and Savarese, Silvio and Hoi, Steven},
  booktitle={ICML},
  year={2023}
}

@article{abdin2024phi,
  title={Phi-3 technical report: A highly capable language model locally on your phone},
  author={Abdin, Marah and Aneja, Jyoti and Awadalla, Hany and Awadallah, Ahmed and Awan, Ammar Ahmad and Bach, Nguyen and Bahree, Amit and Bakhtiari, Arash and Bao, Jianmin and Behl, Harkirat and others},
  journal={arXiv preprint arXiv:2404.14219},
  year={2024}
}

@inproceedings{maaz2023video,
  title={Video-chatgpt: Towards detailed video understanding via large vision and language models},
  author={Maaz, Muhammad and Rasheed, Hanoona and Khan, Salman and Khan, Fahad Shahbaz},
  booktitle={ACL},
  year={2024}
}

@inproceedings{li2024llama,
  title={Llama-vid: An image is worth 2 tokens in large language models},
  author={Li, Yanwei and Wang, Chengyao and Jia, Jiaya},
  booktitle={ECCV},
  year={2024}
}

@inproceedings{bain2021frozen,
  title={Frozen in time: A joint video and image encoder for end-to-end retrieval},
  author={Bain, Max and Nagrani, Arsha and Varol, G{\"u}l and Zisserman, Andrew},
  booktitle={ICCV},
  pages={1728--1738},
  year={2021}
}

@inproceedings{liu2023visual,
  title={Visual Instruction Tuning}, 
  author={Haotian Liu and Chunyuan Li and Qingyang Wu and Yong Jae Lee},
  booktitle={NeurIPS},
  year={2024}
}

@inproceedings{caba2015activitynet,
  title={Activitynet: A large-scale video benchmark for human activity understanding},
  author={Caba Heilbron, Fabian and Escorcia, Victor and Ghanem, Bernard and Carlos Niebles, Juan},
  booktitle={CVPR},
  year={2015}
}

@inproceedings{gao2017tall,
  title={Tall: Temporal activity localization via language query},
  author={Gao, Jiyang and Sun, Chen and Yang, Zhenheng and Nevatia, Ram},
  booktitle={ICCV},
  year={2017}
}

@inproceedings{zhou2018towards,
  title={Towards automatic learning of procedures from web instructional videos},
  author={Zhou, Luowei and Xu, Chenliang and Corso, Jason},
  booktitle={AAAI},
  year={2018}
}

@inproceedings{krishna2017dense,
  title={Dense-captioning events in videos},
  author={Krishna, Ranjay and Hata, Kenji and Ren, Frederic and Fei-Fei, Li and Carlos Niebles, Juan},
  booktitle={ICCV},
  year={2017}
}

@inproceedings{lei2021detecting,
  title={Detecting moments and highlights in videos via natural language queries},
  author={Lei, Jie and Berg, Tamara L and Bansal, Mohit},
  booktitle={NeurIPS},
  year={2021}
}

@inproceedings{vedantam2015cider,
  title={Cider: Consensus-based image description evaluation},
  author={Vedantam, Ramakrishna and Lawrence Zitnick, C and Parikh, Devi},
  booktitle={CVPR},
  year={2015}
}

@inproceedings{fujita2020soda,
  title={Soda: Story oriented dense video captioning evaluation framework},
  author={Fujita, Soichiro and Hirao, Tsutomu and Kamigaito, Hidetaka and Okumura, Manabu and Nagata, Masaaki},
  booktitle={ECCV},
  year={2020}
}

@article{liu2024triadic,
  title={Triadic temporal-semantic alignment for weakly-supervised video moment retrieval},
  author={Liu, Jin and Xie, JiaLong and Zhou, Fengyu and He, Shengfeng},
  journal={Pattern Recognition},
  year={2024}
}

@article{liu2022dynamic,
  title={Dynamic self-attention with vision synchronization networks for video question answering},
  author={Liu, Yun and Zhang, Xiaoming and Huang, Feiran and Shen, Shixun and Tian, Peng and Li, Lang and Li, Zhoujun},
  journal={Pattern Recognition},
  year={2022}
}

@article{chen2026super,
  title={Super encoding network: Recursive association of multi-modal encoders for video understanding},
  author={Chen, Boyu and Chen, Siran and Li, Kunchang and Xu, Qinglin and Qiao, Yu and Wang, Yali},
  journal={Pattern Recognition},
  year={2026}
}

@article{chen2026ask,
  title={Ask and focus more: Question-prompt uncertainty allocation for dual-controllable video captioning},
  author={Chen, Shuqin and Yang, Xingrui and Chen, Yi and Wang, Kai and Yu, Xiaohan and Zhong, Xian},
  journal={Pattern Recognition},
  year={2026}
}

@article{ma2025retta,
  title={Retta: Retrieval-enhanced test-time adaptation for zero-shot video captioning},
  author={Ma, Yunchuan and Qing, Laiyun and Li, Guorong and Qi, Yuankai and Beheshti, Amin and Sheng, Quan Z and Huang, Qingming},
  journal={Pattern Recognition},
  year={2025}
}
